\documentclass[11pt]{article}

\usepackage[final]{acl}

\usepackage{times}
\usepackage{latexsym}

\usepackage[T1]{fontenc}

\usepackage[utf8]{inputenc}

\usepackage{microtype}
\usepackage{multirow} 
\usepackage{inconsolata}
\usepackage{graphicx}

\usepackage{booktabs}   
\usepackage{array}      
\usepackage{longtable}
\usepackage[table]{xcolor}

\usepackage{listings}
\lstdefinestyle{prompt}{
  basicstyle=\footnotesize\ttfamily,
  breaklines=true, breakindent=0pt,
  columns=fullflexible, keepspaces=true,
  upquote=true,                 
  frame=tb, framerule=0.4pt,    
  aboveskip=4pt, belowskip=4pt,
}
\usepackage[table,dvipsnames]{xcolor}  
\usepackage{xfp}                       

\definecolor{heatblue}{HTML}{2A6FB5}   

\usepackage{color}
\usepackage[dvipsnames]{xcolor}  
\usepackage[normalem]{ulem}      

\usepackage{float}
\usepackage{placeins}
\usepackage{enumitem}

\newif\ifauthornotes
\newif\ifstrike
\newif\iftodo
\newif\ifrevise
\newif\ifadd
\newif\ifreplace

\authornotestrue  
\striketrue       

\addtrue      

\title{CliniCIRCA: A Modular LLM Framework for Constructing Longitudinal Mental Health Patient Journeys from Raw EHR Narratives}

\author{
  \textbf{Aiwei Ivy Zhang}\textsuperscript{1} \quad
  \textbf{Nimra Ishfaq}\textsuperscript{2} \quad
  \textbf{Mohit Chandra}\textsuperscript{1} \quad
  \textbf{Santiago Alvarez Lesmes}\textsuperscript{3} \\
  \textbf{Adam Coscia}\textsuperscript{1} \quad
  \textbf{Khatiya Chelidze Moon}\textsuperscript{3} \quad
  \textbf{Xiaohan Ding}\textsuperscript{1} \quad
  \textbf{Munmun De Choudhury}\textsuperscript{1} \\
  \\[0.5em]
  {\normalfont
  \textsuperscript{1}Georgia Institute of Technology \quad
  \textsuperscript{2}University of Texas at Austin \quad
  \textsuperscript{3}Northwell Health}
}

\begin{document}
\maketitle


\begin{abstract}
In mental health care, reasoning over patient journeys is a key task for clinicians.
Yet these journeys, encompassing a longitudinal progression of biological, psychological, and social events, are often spread across disparate unstructured text narratives, making temporal recovery challenging.
We present \textbf{\texttt{CliniCIRCA}}, a multi-stage LLM framework for \textbf{C}alendar-anchored, \textbf{I}mprecision-aware \textbf{R}econstruction of \textbf{C}linical \textbf{A}nnals.
To our knowledge, \texttt{CliniCIRCA} is the first to temporally classify clinical events across unstructured discharge summaries without event-level timestamps. 
From $14,882$ MIMIC-III mental health admissions, we first construct a benchmark of $52$ discharge summaries on which \texttt{CliniCIRCA} produces $15,891$ temporally tagged events.
After correcting $629$ errors based on a clinician-in-the-loop evaluation, we produce verified gold-standard labels.
Finally, the corrected timelines drive a temporally grounded summarization stage that compresses each source $1.52$ times into a date-grouped chronological record.
We then scale the framework to generate $1,000$ silver-standard timelines and evaluate them as training data. Compared with zero- and few-shot prompting, instruction tuning generally improves five open-weight models on event extraction, temporal tagging, and summarization across silver and clinician-verified evaluations.






\end{abstract}

\section{Introduction}
\label{sec:introduction}

Every patient follows a clinical journey: a longitudinal progression of symptoms, diagnoses, treatments, life events, and responses to care that demonstrate how their health has evolved over time \cite{kraljevic_foresightgenerative_2024, ruan_representation_2019}.
Clinicians draw on this progression to interpret a patient's current presentation and guide treatment.

Unfortunately, these journeys are often not documented as explicit timelines \cite{loftus_longitudinal_2024, linhares_clinicalpath_2023, olex_review_2021}. 
For example, discharge summaries are a rich source of longitudinal patient information.
They synthesize prior history, hospital course, and clinical outcomes in free-form text, where events may be repeated, narrated nonsequentially, or linked only through relative temporal expressions and surrounding context \cite{k_clinstructor_2025, seinen_using_2025, lee_prospects_2024, kim_challenges_2024}. 

Recovering temporal structure is thus a central challenge in clinical NLP for modeling patient journeys from clinical text \cite{amirahmadi_trajectory-ordered_2025, makarov_large_2025}.
Patient journeys are often spread across disparate text sources, limiting the use of these narratives by computational systems that require structured longitudinal representations \cite{chaturvedi_temporal_2024, moharasan_extraction_2019}. 
As a result, information fragmentation shifts the burden of reconstruction to clinicians and computational systems.
Clinicians must piece together the patient’s course under time and information constraints, while models must infer temporal relations that remain implicit \cite{linhares_clinicalpath_2023, gao_etal_2022_summarizing, olex_review_2021}.

\begin{figure*}[t]
  \includegraphics[width=\textwidth]{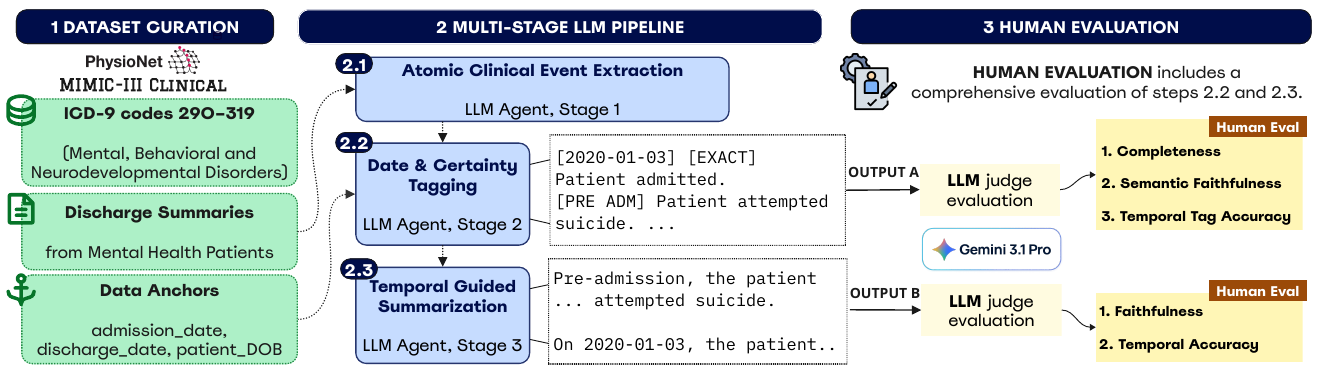}
  \caption{Methodology Overview}
  \label{fig:methodology_overview}
  \vspace{-0.1in}
\end{figure*}

Prior work treated event reconstruction as a set of separable subtasks, which makes temporal alignment challenging. 
For example, clinical summarization compresses lengthy records into coherent accounts, but often under-preserves temporality and rarely represents complex or uncertain event chronology explicitly \cite{Kruse2025Large, cui_timer_2025, croxford_development_2025}. 
Temporal information extraction preserves finer structure by extracting time expressions, events, and temporal relations, but most systems are built around predefined annotation schemas and pairwise relation classification rather than recovery of an open-ended patient trajectory \cite{chaturvedi-etal-2025-temporal, gumiel_temporal_2022, alfattni_extraction_2020}. 

Recent LLM timeline generation comes closer to recovering full patient trajectories \cite{kumar_text_2026, noroozizadeh_temporally_2025, wang_large-language_2025, wang_wonder_2024}, but often relies on structured EHR timestamps, preselected note fragments, or curated case reports, and commonly represents time using a single relative offset. What remains missing is a unified, text-only reconstruction formulation that preserves event breadth, calendar timing, temporal uncertainty, and patient-level coherence. 

We specifically focus on bridging the gap in mental health care, where a patient's course cannot be understood only from diagnoses and procedures alone. Engel's framework emphasizes that illness emerges through the interaction of biological, psychological, and social factors \cite{engel_need_1977}; in mental health, these factors are often expressed through medication use, trauma, social history, changes in daily functioning, etc \cite{saha_life_2026}. Such information is part of the trajectory rather than peripheral context, yet the combination of events differs across patients. Mental health records therefore expose the limitations of fixed schemas: a narrow schema would omit events that may define an individual cause, while an overly exhaustive one quickly becomes impractical. 

Hence, we formulate patient-journey recovery as \textbf{open-vocabulary, calendar-anchored, and uncertainty-aware temporal reconstruction from raw clinical narratives}. Specifically, \textit{journeys are reconstructed within a single discharge summary corresponding to one hospital admission, as a step toward multi-encounter longitudinal modeling}. 
We introduce \textbf{\texttt{CliniCIRCA}} (Figure~\ref{fig:methodology_overview}), a three-stage LLM framework to convert discharge summaries to coherent temporal clinical trajectories. \texttt{CliniCIRCA} produces two linked representations: Output A, an inspectable event-level timeline for computational use and auditing, and Output B, a clinician-readable account derived from the same temporal substrate. Our experiments demonstrate that patient journeys produced by a high-capacity model (Section~\ref{downstream_eval_open_weight_models}) can provide effective supervision for substantially smaller models, particularly for atomic clinical event extraction, while temporal reconstruction remains reasoning-intensive. By transforming fragmented narratives into linked structured and readable representations, \texttt{CliniCIRCA} could support longitudinal chart review, clinical handoffs, and follow-up care while preserving an auditable connection to the underlying events. We have provided code in a repository here\footnote{https://anonymous.4open.science/r/CliniCIRCA-E5FF/}.

\section{Dataset}

We use MIMIC-III v1.4 \cite{johnson_mimic-iii_2016,johnson_mimic-iii_2015}, a publicly available, de-identified critical-care database from the Beth Israel Deaconess Medical Center (2001–2012) in English. \texttt{CliniCIRCA} receives unprocessed discharge summaries without chunking, section parsing, or normalization. Structured EHR fields are used to select the cohort and provide three temporal anchors: (1) date of birth, (2) admission date, and (3) discharge date. We define the cohort at the admission level: an admission is retained if it contains at least one psychiatric diagnosis from the ICD-9 Mental Disorders chapter (codes 290–319) and can be paired with a discharge summary. This produces $14,882$ summaries from $12,273$ patients, each summary treated as a separate reconstruction instance. Because admissions are flagged by the presence of any Mental Disorders diagnosis code rather than by principal diagnosis, the cohort includes both primary psychiatric admissions and medical/surgical admissions with psychiatric comorbidity. More details 
can be found in Appendix~\ref{app:dataset_info}.

\section{The \textbf{\texttt{CliniCIRCA}} Framework}
{\texttt{CliniCIRCA}} is a sequential framework comprising three LLM stages (Figure~\ref{fig:methodology_overview}), each receiving the output of the previous stage and having a single main task. \textbf{Stage 1} reads the raw discharge summary (obtained from the MIMIC-III Dataset) without preprocessing and extracts a high-recall list of atomic clinical events. \textbf{Stage 2} resolves the timing of each event, tagging it with an ISO date and a certainty label anchored against temporal references; its output is \textbf{Output A}, a timeline of (ISO date and certainty tag, atomic event) pairs. \textbf{Stage 3} aggregates these tagged events by date into \textbf{Output B}, a date-grouped chronological summary. Decomposing the task in this way allows each stage to be prompted, evaluated, and improved independently, and confines each temporal decision to the stage best positioned to make it.


\section{Model and Agent Evaluation Setup}
\label{model_agent_eval_setup}
\vspace{-0.2em}
\subsection{Candidate Pool}
For {\texttt{CliniCIRCA}}, we select $11$ LLM candidates based on reported performance across general-purpose and clinical benchmarks \cite{yu_simulated_2025, sarvari_rapidly_2025, shi_ehragent_2024}. For proprietary models, we evaluate Gemini 2.5 Pro and 2.5 Flash \cite{comanici2025gemini25pushingfrontier}; Claude Sonnet 4.6~\cite{anthropic2026sonnet46} and Haiku 4.5~\cite{anthropic2025haiku45}. For open-weight models, we evaluate Llama 3.3 70B~\cite{meta2024llama33}, Qwen 3.6 35B-A3B ~\cite{qwen2026qwen36}, Gemma 4 26B-A4B~\cite{gemmateam2026gemma4}, Mistral Small 3.2 24B-A4B~\cite{mistral2025small32}, Phi-4 Mini 3.8B~\cite{abouelenin2025phi4mini}, and medical-domain LLMs - MedGemma 27B \cite{sellergren2026medgemmatechnicalreport} and Meditron 3 Qwen 2.5 7B \cite{openmeditron2025meditron3}. Model specifications, generation settings, and computational details are provided in Appendix~\ref{app:agent_candidate_selection}.

\subsection{Implementation Setup}
\label{sec:implementation_setup}
We construct a benchmark set of $52$ MIMIC-III discharge summaries, averaging $1,774$ words each (range: $437$–$3,583$; Appendix~\ref{app:benchmark_input_text_profile}). Of the 52 cases, 29 had a principal psychiatric diagnosis; the remaining 23 were medical or surgical admissions with psychiatric comorbidity. For the benchmark, every candidate is run through Stages 1-2, with no mixing of models across stages. 
The size is determined by a power analysis (Appendix~\ref{app:power_analysis}) for a pairwise $\chi^2$ comparison (large effect, $\alpha$ = $0.05$, $95$\% power), yielding \textit{N} = $52$  needed to reach statistical significance \cite{ki2025linguistic, card_little_2020, dror_hitchhikers_2018}.

Each candidate is run through Stages 1-2 as a fixed pipeline configuration, producing $572$ model–document outputs evaluated using the clinician-authored rubric described in Section~\ref{sec:recover_patient_timeline_s1s2}. We retain the selected model for Stage 3 rather than optimizing a separate model for each stage. This avoids a combinatorial search over model combinations, reveals how one model’s capabilities propagate through the {\texttt{CliniCIRCA}}  
framework, and simplifies reproducibility and deployment governance \cite{umeton_gpt-4_2024, de_hond_text_2024, mesko_imperative_2023}. We select based on Stage 2 because the summary stage depends on the accuracy of its event-level temporal input. Following prior health-domain work that validates LLM judges against expert-vetted human labels \cite{mittal_follow_2027}, two human annotators and two clinicians, all with relevant annotation experience and high proficiency in English, form a feedback loop to evaluate the results. We establish human-evaluator agreement on both Stage-2 and 3 outputs.

\section{Constructing Patient Timelines from Discharge Summaries}
\label{sec:recover_patient_timeline}
\subsection{Stage 1 \& 2: From Raw Discharge Summary to Tagged Timeline}
\label{sec:recover_patient_timeline_s1s2}
Faithful clinical timeline recovery requires both comprehensive event extraction and accurate temporal anchoring, yet extraction errors can propagate through temporal reasoning and timeline construction \cite{gumiel_temporal_2022, alfattni_extraction_2020}. We therefore decompose the task into Stage 1 for event extraction and Stage 2 for temporal anchoring, allowing each component to be independently specified and evaluated. Full generation prompts can be found in Appendix~\ref{app:stage1_stage2_gen_prompts} Tables~\ref{tab:appendix_stage1_prompt} -~\ref{tab:appendix_stage2_prompt}.

\begin{table*}[t]
  \centering
  \small
  \setlength{\tabcolsep}{5pt}
  \renewcommand{\arraystretch}{1.15}

  \begin{tabular}{lccccccc}
    \toprule
    \textbf{Model}
    & \textbf{Dimension A}
    & \textbf{Dimension B}
    & \textbf{Dimension C}
    & \textbf{Avg.\ ABC}
    & \textbf{MED-P}
    & \textbf{MED-R}
    & \textbf{MED-F1} \\
    \midrule

    \rowcolor{yellow!15}
    Gemini 2.5 Pro
    & $\textbf{5.00}$
    & $\textbf{5.00}$
    & $\textbf{4.83}$
    & $\textbf{4.94}$
    & $0.637$
    & $\textbf{0.812}$
    & $\textbf{0.710}$ \\

    Claude Haiku 4.5
    & $\textbf{5.00}$
    & $\underline{4.98}$
    & $\underline{4.81}$
    & $\underline{4.93}$
    & $0.610$
    & $0.753$
    & $0.672$ \\

    Claude Sonnet 4.6
    & $\underline{4.98}$
    & $\textbf{5.00}$
    & $4.77$
    & $4.92$
    & $0.614$
    & $0.735$
    & $0.666$ \\

    Gemini 2.5 Flash
    & $\underline{4.98}$
    & $\textbf{5.00}$
    & $4.50$
    & $4.83$
    & $0.617$
    & $\underline{0.762}$
    & $0.679$ \\

    Qwen 3.6 35B-A3B
    & $\textbf{5.00}$
    & $\textbf{5.00}$
    & $4.23$
    & $4.74$
    & $0.630$
    & $0.758$
    & $\underline{0.685}$ \\

    Gemma 4 26B-A4B
    & $4.88$
    & $4.96$
    & $4.27$
    & $4.71$
    & $0.636$
    & $0.713$
    & $0.669$ \\

    Mistral Small 3.2 24B-A4B
    & $4.96$
    & $\underline{4.98}$
    & $3.88$
    & $4.61$
    & $0.636$
    & $0.740$
    & $0.681$ \\

    Llama 3.3 70B
    & $4.77$
    & $4.90$
    & $3.44$
    & $4.37$
    & $0.626$
    & $0.673$
    & $0.644$ \\

    MedGemma 27B
    & $\underline{4.98}$
    & $\underline{4.98}$
    & $2.67$
    & $4.21$
    & $0.638$
    & $0.729$
    & $0.678$\\

    Meditron 3 (Qwen 2.5 7B)
    & $3.71$
    & $4.73$
    & $2.00$
    & $3.48$
    & $\underline{0.680}$
    & $0.285$
    & $0.361$ \\

    Phi-4 Mini 3.8B
    & $3.90$
    & $4.17$
    & $2.19$
    & $3.42$
    & $\textbf{0.685}$
    & $0.461$
    & $0.525$ \\

    \bottomrule
  \end{tabular}

  \vspace{1pt}
    
  \caption{Stage-2 model-selection results on the 52-sample benchmark.
  The highlighted row indicates the model selected for the final pipeline.
  Results are averaged over 52 samples.
  Dimension A: Completeness;
  Dimension B: Semantic Faithfulness; and
  Dimension C: Temporal Tag Accuracy.
  Dimensions A to C are Likert-scale averages reported to two decimal places.
  The best-performing value for each metric is shown in \textbf{bold}, and the second-best value is \underline{underlined}.
  Tied values receive the same formatting.
  }
  \label{tab:model_selection}
  \vspace{-0.1in}
\end{table*}

\textbf{Inclusive extraction and anchor-based tagging.} Stage 1 reads the raw discharge summary, applying no chunking, section parsing, or normalization, and extracts events inclusively and faithfully.
Accordingly, Stage 2 then resolves each event against the admission date, discharge date, and date of birth, assigning exactly one of four temporal categories: 1)\texttt{{[DATE/RANGE] [EXACT]}}: a date written in the event text itself. 2) \texttt{{[DATE/RANGE] [APPROX]}}: no written date, but the event resolves to an anchor or a relative offset from one. 3) \texttt{{[PRE ADM]}}: framed as prior to or continuous through this admission (history, “status post,” demographics, undated chronic states), and 4) \texttt{{[INDETERMINATE]}}: a guardrail used only when none of the above applies and the timing genuinely cannot be determined. 

As specified above, to maximize precision, only explicitly written dates receive {[EXACT]}; any inferred or relative date resolves to {[APPROX]} at the coarsest defensible granularity (e.g., year-only when only a year is recoverable). Stage 2's output is a list of temporally classified clinical events.

\paragraph{Operationalizing the notion of “faithfulness”.} Because no reference timeline exists against which these tuples could be scored, a “good” timeline must itself be operationalized. We define three clinician-informed criteria for evaluating the Stage-2 output (full definitions in Appendix~\ref{app:stage1_stage2_eval_rubrics}).\\
\textbf{A) Completeness}: \textit{whether all clinically relevant events in the note are captured.}\\
\textbf{B) Semantic faithfulness}: \textit{whether each extracted event preserves the meaning of the source text.}\\
\textbf{C) Temporal classification accuracy}: \textit{whether the assigned date and tag are correct.}

\paragraph{Calibrating the automated judge.} Following established practice in LLM evaluation to use a strong later-generation model \cite{zheng_2023_llm_judge}, we select Gemini 3.1 Pro as the LLM evaluator, chosen to be at least as capable as any candidate it scores. Prior to selection, the evaluator is calibrated against expert human scoring. Only the calibrated LLM evaluator is used in the selection and scaling.

\textbf{Selecting the agent model.} Following the agent selection mechanism as described in Section~\ref{sec:implementation_setup}, Gemini 2.5 Pro is selected as the strongest configuration, ranking among the top candidates on every dimension and achieving $5.00$ on Completeness, $5.00$ on Semantic Faithfulness, and $4.83$ on Temporal Tag Accuracy in Table~\ref{tab:model_selection}. As a judge-independent check, we also report MEDCON \cite{aci-bench}, the F1 overlap of UMLS medical concepts between each model's extracted events and the source note; once again, Gemini 2.5 Pro attains the highest MEDCON F1 (Table~\ref{tab:model_selection}), corroborating its judge-scored ranking. We select based on Stage 2 alone and then freeze Gemini 2.5 Pro as the only agent advancing to Stage 3. 

\subsection{Validating the Stage 1-2 Timeline}
\label{sec:stage1-2_validation}
We validate the output of Stage 1-2 through a human review with two psychiatric clinicians during 6 iterative annotation sessions. 
The human review process was used both to evaluate the automated judge and to correct all confirmed errors, providing a clinician-verified gold reference dataset ($N=52$). Evaluation was conducted on an internally hosted, custom web-based secure platform, demonstrated in Appendix~\ref{app:stage1_stage2_eval_platform}.

We compare the calibrated Gemini 3.1 Pro evaluator with the clinician-verified gold annotation (see Appendix~\ref{app:stage1_stage2_eval_rubrics} for complete rubrics). Overall agreement is high, with Cohen's $\kappa=0.655$. However, because only $3.96\%$ of events were labeled as errors, class imbalance likely depresses $\kappa$ \cite{gwet2008computing}. This is reflected in a Gwet's AC1 of $0.972$ and negative agreement of $0.986$, indicating strong agreement on events judged correct. Agreement is lower for the rare error class, with positive agreement of $0.669$.



\subsection{Stage 3: From Timeline to Temporally-Guided Summary}
Next, Stage 3 tests if the verified event timeline can be rendered as a faithful, temporally grounded summary without re-reading the discharge note or reconstructing its chronology. Two clinicians defined a specification for the mental-health cohort that retains demographics and social history, reason for admission, key in-stay events, medication changes, discharge, and follow-up. Each event is grouped under its Stage 2 temporal label.

Stage 3 is evaluated for faithfulness to the verified timeline and temporal accuracy using a separately calibrated Gemini 3.1 Pro evaluator validated against clinician judgments. Generation prompts and rubrics appear in Appendices~\ref{app:stage3_gen_prompts} -~\ref{app:stage3_eval_rubrics}. Across the benchmark, Stage 3 reduces reading burden by roughly a third (31.5\% mean and 33.8\% median per-note; 34.0\% corpus-wide), a stable 1.52 times compression that removes $\sim$600 words per admission (Table~\ref{tab:stage3_reduction}).

\begin{table}[t]
\centering
\small
\begin{tabular}{lr}
\toprule
\textbf{Metric} & \textbf{Value} \\
\midrule
Mean reduction (avg.\ per-note \%)     & 31.53\% \\
Median per-note reduction             & 33.76\% \\
Corpus-level reduction (total words)  & 34.01\% \\
Compression factor                    & 1.52$\times$ \\
\midrule
Mean words reduced per admission      & 603 \\
Median words reduced                  & 422 \\
\bottomrule
\end{tabular}
\caption{Reading-burden reduction from Stage 3 summarization over the 52-note benchmark. Per-note percentages are averaged across notes; corpus-level reduction pools word counts across all notes.}
\label{tab:stage3_reduction}
\vspace{-0.1in}
\end{table}

\subsection{Validating the Stage 3 Summary}
\label{sec:stage3_validation}
We compare the judge's error counts against human adjudication 
across $52$ summaries (Table~\ref{tab:stage3_error_count_concordance}). The judge flags $262$ errors, of which human annotators uphold $39$: a $6.7$-fold inflation, averaging $4.3$ spurious errors per summary. Agreement is substantial on whether a summary contains any error ($\kappa = 0.77$, positive agreement = 0.84) and on the relative ordering of summaries by error count (Spearman $\rho$ = $0.77$). With clinician guidance, we develop a two-dimensional rubric (Tables~\ref{tab:appendix_stage3_faithfulness}-\ref{tab:appendix_stage3_temporal}) to validate if generated summaries: (A) faithfully retained relevant mental-health and social events without distortion and (B) placed them accurately in time. See Appendix~\ref{app:stage3_eval_agreement} for full metrics and results, and Appendix~\ref{app:stage3_eval_platform} for the internally hosted, custom web-based evaluation platform.

\subsection{Qualitative Analysis}
\label{sec:qualitative_analysis}
We qualitatively investigate what this reconstructed timeline helps a reader (e.g., a clinician) understand. To examine the clinical content captured by the timelines, we analyze three clinician-verified gold cases using two World Health Organization frameworks: eight categories of mental disorders and associated symptoms \cite{who_mental_disorders_2025}, and three levels of risk factors: Individual, Family \& Community, and Structural \cite{who_mental_health_2025}. 
We label a risk factor as co-documented rather than causal unless the record explicitly links it to the disorder. Across three exemplar cases, temporal reconstruction distinguishes long-standing risk factors from acute in-stay changes and diagnostic reassessment, while preserving whether their relationships are causal, co-documented, or unresolved rather than collapsing them into a single causal narrative. Full case analyses are provided in Appendix~\ref{app:qualitative_analysis_cases}.



\section{Evaluation with Open-Weight Models}
\label{downstream_eval_open_weight_models}
\begin{table*}[!t]
\centering
\small
\setlength{\tabcolsep}{4.5pt}
\renewcommand{\arraystretch}{1.15}
\begin{tabular}{@{}lccccccccc@{}}
\toprule
& \multicolumn{3}{c}{\textsc{Silver$\rightarrow$Silver}}
& \multicolumn{3}{c}{\textsc{Silver$\rightarrow$Gold}}
& \multicolumn{3}{c}{\textsc{Gold}} \\
\cmidrule(lr){2-4} \cmidrule(lr){5-7} \cmidrule(lr){8-10}
\textbf{Model}
& \textbf{Zero} & \textbf{Few} & \textbf{IT}
& \textbf{Zero} & \textbf{Few} & \textbf{IT}
& \textbf{Zero} & \textbf{Few} & \textbf{IT} \\
\midrule

Gemma4-31B
& \cellcolor{blue!20}$.640_{.011}$ & \cellcolor{blue!20}$.665_{.016}$ & \cellcolor{blue!30}$.840_{.008}$
& \cellcolor{blue!20}$.639_{.012}$ & \cellcolor{blue!20}$.659_{.017}$ & \cellcolor{blue!30}$.827_{.014}$
& -- & \cellcolor{blue!25}$.727_{.013}$ & \cellcolor{blue!20}$.628_{.016}$ \\

Qwen3.6-35B-A3B
& \cellcolor{blue!16}$.536_{.014}$ & \cellcolor{blue!20}$.642_{.013}$ & \cellcolor{blue!30}$.839_{.009}$
& \cellcolor{blue!16}$.535_{.021}$ & \cellcolor{blue!20}$.647_{.017}$ & \cellcolor{blue!30}$.834_{.012}$
& -- & \cellcolor{blue!25}$\mathbf{.732}_{.012}$ & \cellcolor{blue!16}$.550_{.026}$ \\

Qwen3-30B-A3B
& \cellcolor{blue!13}$.420_{.027}$ & \cellcolor{blue!16}$.547_{.025}$ & \cellcolor{blue!25}$.745_{.028}$
& \cellcolor{blue!13}$.454_{.028}$ & \cellcolor{blue!16}$.561_{.020}$ & \cellcolor{blue!25}$.795_{.013}$
& -- & \cellcolor{blue!8}$.279_{.045}$ & \cellcolor{blue!20}$.606_{.033}$ \\

Mistral-24B
& \cellcolor{blue!10}$.309_{.021}$ & \cellcolor{blue!16}$.565_{.016}$ & \cellcolor{blue!30}$\mathbf{.848}_{.008}$
& \cellcolor{blue!8}$.284_{.021}$ & \cellcolor{blue!16}$.545_{.032}$ & \cellcolor{blue!30}$\mathbf{.847}_{.009}$
& -- & \cellcolor{blue!20}$.676_{.015}$ & \cellcolor{blue!25}$.715_{.028}$ \\

Meditron-7B
& \cellcolor{blue!3}$.040_{.006}$ & \cellcolor{blue!5}$.131_{.013}$ & \cellcolor{blue!25}$.744_{.021}$
& \cellcolor{blue!3}$.032_{.007}$ & \cellcolor{blue!5}$.120_{.015}$ & \cellcolor{blue!25}$.752_{.021}$
& -- & \cellcolor{blue!3}$.034_{.004}$ & \cellcolor{blue!10}$.331_{.024}$ \\

\bottomrule
\end{tabular}
\caption{Task~A measured by embedding-based
RM-F1@0.85. \textsc{Silver$\rightarrow$Silver}
trains and evaluates on the silver split;
\textsc{Silver$\rightarrow$Gold} trains on silver and evaluates on all
$52$ clinician-verified examples; \textsc{Gold} uses four-fold
cross-validation on the gold set alone. Results are point estimates $\pm$ bootstrap SD.} 
\label{tab:task-a-results}
\vspace{-.1in}
\end{table*}



\begin{table*}[!t]
\centering

\small
\setlength{\tabcolsep}{3.7pt}
\renewcommand{\arraystretch}{1.10}

\begin{tabular}{@{}lccccccccc@{}}
\toprule
& \multicolumn{3}{c}{\textsc{Silver$\rightarrow$Silver}}
& \multicolumn{3}{c}{\textsc{Silver$\rightarrow$Gold}}
& \multicolumn{3}{c}{\textsc{Gold}} \\
\cmidrule(lr){2-4}
\cmidrule(lr){5-7}
\cmidrule(lr){8-10}

\textbf{Model}
& \textbf{Zero} & \textbf{Few} & \textbf{IT}
& \textbf{Zero} & \textbf{Few} & \textbf{IT}
& \textbf{Zero} & \textbf{Few} & \textbf{IT} \\
\midrule

Gemma4-31B
& \cellcolor{blue!25}$.738_{.019}$ & \cellcolor{blue!25}$.731_{.019}$ & \cellcolor{blue!25}$\mathbf{.767}_{.024}$
& \cellcolor{blue!25}$.799_{.036}$ & \cellcolor{blue!30}$\mathbf{.804}_{.043}$ & \cellcolor{blue!25}$.749_{.029}$
& -- & \cellcolor{blue!20}$.668_{.029}$ & \cellcolor{blue!25}$\mathbf{.721}_{.035}$ \\

Qwen3.6-35B-A3B
& \cellcolor{blue!20}$.635_{.021}$ & \cellcolor{blue!20}$.687_{.020}$ & \cellcolor{blue!25}$.742_{.019}$
& \cellcolor{blue!16}$.567_{.020}$ & \cellcolor{blue!25}$.715_{.025}$ & \cellcolor{blue!25}$.731_{.022}$
& -- & \cellcolor{blue!20}$.602_{.023}$ & \cellcolor{blue!20}$.633_{.019}$ \\

Qwen3-30B-A3B
& \cellcolor{blue!13}$.426_{.017}$ & \cellcolor{blue!20}$.609_{.021}$ & \cellcolor{blue!20}$.695_{.016}$
& \cellcolor{blue!13}$.452_{.019}$ & \cellcolor{blue!20}$.628_{.031}$ & \cellcolor{blue!20}$.658_{.011}$
& -- & \cellcolor{blue!8}$.231_{.038}$ & \cellcolor{blue!13}$.425_{.020}$ \\

Mistral-24B
& \cellcolor{blue!16}$.573_{.013}$ & \cellcolor{blue!20}$.638_{.024}$ & \cellcolor{blue!25}$.733_{.025}$
& \cellcolor{blue!16}$.567_{.040}$ & \cellcolor{blue!20}$.615_{.023}$ & \cellcolor{blue!25}$.734_{.018}$
& -- & \cellcolor{blue!20}$.620_{.012}$ & \cellcolor{blue!20}$.680_{.007}$ \\

Meditron-7B
& \cellcolor{blue!8}$.266_{.016}$ & \cellcolor{blue!3}$.004_{.003}$ & \cellcolor{blue!13}$.401_{.035}$
& \cellcolor{blue!8}$.256_{.018}$ & \cellcolor{blue!3}$.004_{.004}$ & \cellcolor{blue!20}$.666_{.048}$
& -- & \cellcolor{blue!3}$.005_{.004}$ & \cellcolor{blue!10}$.359_{.022}$ \\

\bottomrule

\end{tabular}
\caption{%
Task~B temporal tagging, scored by macro-averaged $\mathrm{F}1$ over
the four temporal tags with the atomic events
supplied as oracle input. Results are point estimates $\pm$ bootstrap SD.%
}
\label{tab:task-b-results}
\vspace{-.1in}
\end{table*}

\begin{table*}[!t]
\centering

\small
\setlength{\tabcolsep}{3.7pt}
\renewcommand{\arraystretch}{1.10}

\begin{tabular}{@{}lccccccccc@{}}
\toprule
& \multicolumn{3}{c}{\textsc{Silver$\rightarrow$Silver}}
& \multicolumn{3}{c}{\textsc{Silver$\rightarrow$Gold}}
& \multicolumn{3}{c}{\textsc{Gold}} \\
\cmidrule(lr){2-4}
\cmidrule(lr){5-7}
\cmidrule(lr){8-10}

\textbf{Model}
& \textbf{Zero} & \textbf{Few} & \textbf{IT}
& \textbf{Zero} & \textbf{Few} & \textbf{IT}
& \textbf{Zero} & \textbf{Few} & \textbf{IT} \\
\midrule

Gemma4-31B
& \cellcolor{blue!20}$.671_{.011}$ & \cellcolor{blue!20}$.647_{.011}$ & \cellcolor{blue!25}$\mathbf{.792}_{.010}$
& \cellcolor{blue!20}$.636_{.015}$ & \cellcolor{blue!20}$.616_{.019}$ & \cellcolor{blue!25}$.749_{.017}$
& -- & \cellcolor{blue!16}$.570_{.026}$ & \cellcolor{blue!20}$.609_{.028}$ \\

Qwen3.6-35B-A3B
& \cellcolor{blue!16}$.538_{.011}$ & \cellcolor{blue!16}$.584_{.011}$ & \cellcolor{blue!25}$.779_{.010}$
& \cellcolor{blue!13}$.490_{.016}$ & \cellcolor{blue!16}$.540_{.017}$ & \cellcolor{blue!25}$\mathbf{.754}_{.015}$
& -- & \cellcolor{blue!13}$.497_{.029}$ & \cellcolor{blue!16}$.516_{.032}$ \\

Qwen3-30B-A3B
& \cellcolor{blue!16}$.504_{.012}$ & \cellcolor{blue!16}$.532_{.012}$ & \cellcolor{blue!25}$.705_{.014}$
& \cellcolor{blue!13}$.474_{.015}$ & \cellcolor{blue!16}$.522_{.015}$ & \cellcolor{blue!20}$.674_{.021}$
& -- & \cellcolor{blue!5}$.108_{.032}$ & \cellcolor{blue!8}$.214_{.030}$ \\

Mistral-24B
& \cellcolor{blue!16}$.513_{.012}$ & \cellcolor{blue!16}$.538_{.012}$ & \cellcolor{blue!25}$.770_{.011}$
& \cellcolor{blue!13}$.450_{.019}$ & \cellcolor{blue!13}$.486_{.017}$ & \cellcolor{blue!25}$.738_{.017}$
& -- & \cellcolor{blue!13}$.484_{.019}$ & \cellcolor{blue!20}$\mathbf{.616}_{.025}$ \\

Meditron-7B
& \cellcolor{blue!10}$.361_{.016}$ & \cellcolor{blue!10}$.328_{.023}$ & \cellcolor{blue!20}$.690_{.018}$
& \cellcolor{blue!8}$.293_{.021}$ & \cellcolor{blue!8}$.244_{.031}$ & \cellcolor{blue!20}$.663_{.021}$
& -- & \cellcolor{blue!5}$.101_{.018}$ & \cellcolor{blue!5}$.159_{.020}$ \\
\bottomrule

\end{tabular}
\caption{
Task~C timeline summarization measured by ROUGE-L.
Models receive temporally tagged events and the source discharge summary as oracle input.
Results are point estimates $\pm$ bootstrap SD.
}
\label{tab:task-c-results}
\vspace{-0.15in}
\end{table*}

We evaluate whether \texttt{CliniCIRCA}’s intermediate outputs can serve as reusable supervision under three training and evaluation settings.

\textsc{Silver$\rightarrow$Silver} measures how well a model fits pipeline-generated supervision. We sample $1{,}000$ admissions disjoint from the $52$ gold examples, label them with the frozen Stage~1--2 pipeline, and split them 
into $750$ training, $100$ validation, and $150$
held-out test documents. This Stage 1 to 3 generation processed $61.84$M input and output tokens in total, averaging $61,836$ tokens per admission across the complete framework (Appendix~\ref{app:token_usage_silver}).

\textsc{Silver$\rightarrow$Gold} measures whether that supervision transfers
to human-verified annotations: models trained on the same $750$ silver
documents are evaluated on all $52$ gold examples.

\textsc{Gold} measures what a small clinician-verified set alone can support.
Because $52$ examples cannot sustain a stable held-out split, we run
four-fold cross-validation and pool the out-of-fold predictions before computing metrics and bootstrap
intervals. Throughout, the gold examples are excluded from silver labeling, silver-trained adapters, and few-shot exemplar selection to prevent unintentional leakage. Under the \textsc{Gold} setting, each document is evaluated only by the fold-specific model for which it was held out. 

\subsection{Experimental Setup}

\paragraph{Models.}
We evaluate five locally deployable open-weight models spanning dense and mixture-of-experts architectures, $7$B--$35$B parameter scales, and general-purpose versus clinical specialization: Gemma4-31B~\cite{gemmateam2026gemma4}, Mistral-24B~\cite{mistral2025small32}, Qwen3.6-35B-A3B~\cite{qwen2026qwen36}, Qwen3-30B-A3B~\cite{qwen3technicalreport}, and Meditron-7B~\cite{openmeditron2025meditron3}. Qwen3-30B-A3B and Gemma4-31B were not evaluated during Stage~2 model selection, allowing us to test whether observed performance extends beyond the models previously selected.

\paragraph{Prompting and tuning.}
For each task, we compare three settings under identical task prompts:
zero-shot prompting, few-shot prompting with three training exemplars,
and LoRA-based instruction tuning with one adapter per model and task.
Instruction tuning is intended to align models to the expected output
format and decision rules rather than to inject new clinical knowledge.

\paragraph{Metrics.}
We evaluate Task~A using relaxed-match F1 with a cosine similarity threshold of $0.85$ (RM-F1@0.85) based on prior research~\cite{sharif2025regen, zhang2019bertscore}. Task~B uses macro-F1 over four temporal categories, with a prediction considered correct when both the temporal label and resolved ISO date or range match the reference. Task~C is evaluated using ROUGE-L; ROUGE-1 and ROUGE-2 are reported in Appendix~\ref{app:additional-results}.

\paragraph{Statistical comparison.}
We report point estimates on the full evaluation set with standard deviations from $2{,}000$ document-level bootstrap resamples. For comparisons on the same evaluation set, resamples are paired across systems, and differences are assessed using $95\%$ confidence intervals for the paired score difference. We describe a difference as reliable when its confidence interval excludes zero; otherwise, we make no claim that one system outperforms the other.

\subsection{Results}

We examine if silver supervision improves open-weight models, transfers to clinician-verified annotations, and provides greater utility than the gold set alone. Note that zero-shot prompting requires no training fold, so its GOLD-condition performance is identical to the zero-shot SILVER$\rightarrow$GOLD condition; we therefore omit the redundant GOLD zero-shot column in Tables \ref{tab:task-a-results}-\ref{tab:task-c-results}. Unless otherwise noted, comparisons below describe point-estimate rankings; we flag a difference as statistically reliable only where the paired bootstrap confidence interval excludes zero, and treat other orderings as suggestive rather than established.

\paragraph{Task A: Atomic event extraction.}
Under \textsc{Silver$\rightarrow$Gold}, instruction tuning yields higher RM-F1@0.85 scores than zero-shot and few-shot prompting for all five models (Table~\ref{tab:task-a-results}). Mistral-24B achieves the highest point estimate at $0.847$, compared with $0.284$ zero-shot and $0.545$ few-shot. Silver-trained models also retain similar point estimates when evaluation shifts from the silver test set to clinician-verified annotations. For example, Mistral-24B scores $0.848$ under \textsc{Silver$\rightarrow$Silver} and $0.847$ under \textsc{Silver$\rightarrow$Gold}. These results show that benefits of silver supervision extend beyond evaluation against the generated labels.

\paragraph{Task B: Temporal tagging.}
Temporal tagging shows a less uniform effect of instruction tuning (Table~\ref{tab:task-b-results}). Under \textsc{Silver$\rightarrow$Gold}, instruction tuning yields higher scores than both prompting baselines for four of the five models. Mistral-24B and Qwen3.6-35B-A3B reach $0.734$ and $0.731$, respectively. Gemma4-31B is the exception, achieving its highest score with few-shot prompting ($0.804$), compared with $0.799$ zero-shot and $0.749$ after instruction tuning.

\paragraph{Task C: Timeline summarization.}
Under \textsc{Silver$\rightarrow$Gold}, instruction tuning yields higher ROUGE-L scores than both prompting baselines for all five models (Table~\ref{tab:task-c-results}). Qwen3.6-35B-A3B achieves the highest point estimate at $0.754$, followed by Gemma4-31B at $0.749$ and Mistral-24B at $0.738$. Meditron-7B shows the largest increase, rising from $0.293$ zero-shot and $0.244$ few-shot to $0.663$ after instruction tuning.


\paragraph{Silver supervision is more effective than gold-only tuning at the current scale.}
Across all three tasks and all five models, instruction tuning on the silver dataset yields higher gold-set point estimates than training on the gold folds alone. For Task~A, Mistral-24B achieves $0.847$ under \textsc{Silver$\rightarrow$Gold}, compared with $0.715$ under \textsc{Gold}; Qwen3-30B-A3B scores $0.795$ and $0.606$, respectively. The same pattern appears in Task~B.

The gold-only results should not be interpreted as evidence that the silver annotations are of higher quality. Rather, they show that at the evaluated scale, broader coverage and larger volume of the silver dataset provide sufficient supervision for downstream models to learn the target tasks. Gold annotations remain essential as an independent benchmark for assessing whether models trained on silver supervision generalize to human-verified outputs.

\section{Related Work}
\subsection{Temporal Annotation in Clinical NLP}
Clinical temporal NLP has traditionally focused on extracting typed events, time expressions, and pairwise temporal relations in benchmarks such as i2b2, THYME, and n2c2 \cite{styler_temporal_2014, sun_evaluating_2013, uzuner_2010_2011}. Later work extended this paradigm toward absolute event timing and temporal intervals \cite{leeuwenberg_towards_2020, yang_using_2024}. These methods established the foundations of clinical temporal reasoning, but generally assume predefined event types or annotated mentions \cite{wang_wonder_2024}. \texttt{CliniCIRCA} instead reconstructs an open-vocabulary, patient-level timeline directly from narrative text, grounding events to calendar dates or ranges while preserving temporal uncertainty.

\subsection{Constructing Event Timelines \& Summaries from Clinical Narratives}
\label{sec:related_work_constructing}
Recent work has used LLMs to generate event time sequences from clinical narratives through either single-pass extraction or multi-stage reconstruction \cite{kumar_text_2026,noroozizadeh_temporally_2025,wang_large-language_2025}. Existing approaches make this task tractable by imposing different forms of structure. MedTimeline focuses on a predefined schema of chemotherapy events and does not model temporally vague events \cite{wang_wonder_2024}. \citet{yang_using_2024} and \citet{kumar_text_2026} rely on structured EHR data to improve or calibrate event timing. MIMIC-IV-Ext-22MCTS applies chunking and retrieval to discharge summaries before generating events with relative timestamps \cite{wang_mimic-iv-ext-22mcts_2025}. In contrast, \citet{noroozizadeh_temporally_2025} construct timelines from published case reports, which are typically more organized and curated than routine clinical narratives. Together, these studies demonstrate the feasibility and downstream value of narrative timeline generation, but they focus on settings that are narrower, more structured, or more curated than ours.


Clinical summarization addresses a complementary goal: condensing lengthy records into coherent accounts for review and decision-making \cite{luo_towards_2023, lee_prospects_2024, croxford_development_2025}. Most summarization systems generate directly from the source note, combining event selection, temporal reasoning, and narrative generation in a single step that makes the underlying chronology difficult to inspect. \texttt{CliniCIRCA} instead decouples these decisions: it first constructs an event-level, uncertainty-aware timeline and then summarizes from that representation. The summary is therefore not an independent compression of the original note, but a readable realization of an auditable temporal substrate.

\section{Broader Implications}
Beyond demonstrating the feasibility of patient-journey construction from discharge summaries, our findings reveal practical implications for clinical NLP and usage in real-world contexts.

\paragraph{Reusable supervision.} Clinical institutions often have abundant unstructured records but limited resources for expert manual annotation, secure computing resources, and access to large proprietary models \cite{balasubramanian_leveraging_2025, ntinopoulos_large_2025, grothey_comprehensive_2025}. \texttt{CliniCIRCA} offers a practical route from raw text to task-specific supervision: instruction tuning on only $1{,}000$ silver-labeled examples improved every model on extraction and summarization, and four of five on temporal tagging. These results suggest that modest pipeline-generated datasets can support smaller, locally deployable models without requiring expert labeling of an entire clinical corpus or complex schema development, as has been the case in prior research (Section~\ref{sec:related_work_constructing}).

\vspace{-0.2em}
\paragraph{The necessity of human-in-the-loop oversight.} Reliable evaluation is particularly important in clinical NLP because infrequent errors can remain consequential even when aggregate performance appears high \cite{jiang_beyond_2026, hager_evaluation_2024}. Prior work in psychiatric NLP similarly finds that LLM outputs may resemble expert responses on surface-level properties such as tone while diverging on clinically consequential dimensions \cite{chandra_lived_2025, wang-etal-2025-feel}. Our automated judge agrees strongly with mental health clinicians on the large majority of correct events, but is less reliable in identifying the nature of rare errors, particularly those involving pre-admission medications and date granularity. Producing a trustworthy reference set therefore requires exhaustive clinician-guided adjudication, since expert review remains necessary to resolve ambiguous temporal decisions and determine which errors are clinically meaningful~\cite{bavaresco-etal-2025-llms, diekmann_llms_2025, chen_humans_2024}. 


\vspace{-0.2em}
\paragraph{Supporting longitudinal chart review.} Reviewing longitudinal records is time- and effort-intensive, especially for multimorbid patients whose histories could be distributed across multiple clinical encounters and narrative documents \cite{van_veen_adapted_2024}. Our proposed Stage 3 renders the event-level timeline as a readable patient journey while retaining its calendar anchors and temporal uncertainty. Such summaries could provide clinicians with a traceable overview before time-constrained follow-up visits, especially in contexts in which manually reconstructing a complex history from several discharge summaries may be impractical. That said, we underscore that they should complement rather than replace the source record, with individual claims remaining connected to inspectable event-level evidence. 

\section{Conclusion and Future Work}
\texttt{CliniCIRCA} shows that recovering a patient journey from clinical narrative is best treated not as a single generation problem, but as separable representational decisions: what happened, when it happened, and how it should be rendered for review. This decomposition makes the resulting chronology inspectable, exposes where temporal ambiguity remains unresolved, and enables each stage to be evaluated independently. Our evaluation further shows that automated judges are useful for locating problematic outputs, but not for replacing expert adjudication of rare temporal errors. At the same time, the reconstructed journeys provide effective supervision for smaller open-weight models, suggesting that structured intermediate representations can serve not only as outputs, but also as reusable training resources. Future work will extend \texttt{CliniCIRCA} to multi-document patient records, richer temporal representations, and additional note types, to support traceable longitudinal chart review and memory-aware clinical systems.


\section*{Limitations}

Although we present a framework with clinician-designed and verified definitions and steps, we recognize several areas for improvement.

\textbf{Representational scope. } Our date-centered schema supports chronological ordering but may impose artificial timing on atemporal facts or persistent states, such as sex, family history, and long-term substance use. Future work should compare alternatives, including atemporal labels, validity intervals, and separate state/event categories, and evaluate their effects on timeline fidelity, summarization, and borderline cases.

\textbf{Summarization salience.} Stage 3 follows a clinician-authored specification for retaining, combining, and omitting events in the narrative summary. Because the same two clinicians developed this specification and conducted the faithfulness and temporal-accuracy evaluation described in Section~\ref{sec:stage3_validation}, the results may understate disagreement in salience or interpretation. Independent evaluation would strengthen this validation. The specification also reflects one view of clinical relevance, since priorities may differ across clinicians and care settings. The resulting summaries should therefore not be treated as a single canonical account of a patient’s journey. Future work should include more diverse evaluators and clinical settings, salience policies, and multiple reference summaries.

\textbf{Psychiatric symptom representation in Stage 3.} Our Stage 3 specification (Appendix~\ref{app:stage3_gen_prompts}) excludes affect descriptions with vital signs and routine examination findings to reduce boilerplate. As a result, mood and affect trajectories -- a core component of psychiatric symptom tracking, as illustrated by the depression case in Section~\ref{sec:qualitative_analysis} -- are preserved in the auditable Output A timeline (e.g., via diagnosis and treatment events) but not always carried into the reader-facing Output B narrative in their own right. Revisiting this exclusion for the mental-health setting specifically, for example by retaining clinically significant affect changes as a distinct category rather than folding them under general exam findings, is a natural next step for future research.

\textbf{Single-document scope.} \texttt{CliniCIRCA} reconstructs the course described within one discharge summary per admission rather than multiple notes from the full longitudinal record. This bounded setting isolates narrative reconstruction, but cannot reconcile duplicated, conflicting, or evolving information across admissions. Prior patient-journey methods are not directly comparable because they use different inputs, prompts, and context assumptions. We therefore compare 11 candidate models under the same framework (Section~\ref{model_agent_eval_setup}), but do not include a single-pass end-to-end baseline. Future work should ablate the modular design using the same LLM and extend it to multi-document records through cross-note entity resolution, temporal alignment, and representation of revisions in clinical understanding.

Despite these limitations, our proposed framework \texttt{CliniCIRCA} provides a clinician-validated representation for open-vocabulary temporal reconstruction from raw clinical narratives.

\section*{Ethical Considerations}
We use the MIMIC-III Clinical Database (v1.4), accessed through PhysioNet under its Credentialed Health Data Use Agreement (v1.5.0) with the required CITI training completed for all personnel with direct access or analysis of the data. LLMs were used in compliance with PhysioNet's policy on responsible use of MIMIC data with LLMs and online services. We accessed Claude models through Amazon Bedrock and Gemini models through Vertex AI because the data were neither externally retained nor used for model training. All human evaluation was performed exclusively by the authors, without the recruitment of external human subjects.
As a secondary analysis of de-identified records, this work is not human-subjects research and required no additional IRB review.

 Because source notes may unevenly document social history, trauma, or substance use across patients -- for instance by diagnosis, demographic group, or documentation era -- \texttt{CliniCIRCA}'s extraction and summarization stages could reproduce such documentation asymmetries rather than correct them; auditing this is an important direction for future work. We also note that pipeline errors are not equally consequential: a mis-tagged or fabricated event involving suicidality, self-harm, or medication timing carries substantially more clinical risk than a mis-tagged laboratory value. Given this, we emphasize that Stage 3 outputs (Output B) are intended to complement, not replace, the source record, and should remain linked to their inspectable Stage 1-2 source events (Output A) in any deployment setting within the healthcare system, particularly for safety-relevant content.

\section*{Acknowledgments}

Zhang, Ding, and De Choudhury were partly supported through funds from The Children's Healthcare of Atlanta Pediatric Technology Center at Georgia Tech. We thank Jiawei Zhou, Rijul Magu, Shravika Mittal, Viet Cuong Nguyen, Pinxian Lu, Owen Xingjian Zhang, Zeyu Hua, and Zikang Leng for valuable technical ideation and suggestions throughout this work. We are grateful to Franklin Ye Ruan for discussions of clinical and medical knowledge that informed the design of this study. We thank Sonakshi Sisodiya for early-stage annotation, and Andrew Zhao for technical and server support. We also thank Kaike Ping for statistical guidance and support.


\bibliography{bib_files/clinicirca_bib, bib_files/new_intro, bib_files/broader_implications, bib_files/models_bib}

\clearpage

\appendix

\renewcommand{\thefigure}{A\arabic{figure}}
\renewcommand{\thetable}{A\arabic{table}}
\setcounter{figure}{0}
\setcounter{table}{0}

\section{Dataset Extraction and Profile}
\label{app:dataset_info}
\subsection{Mental Health Admissions Subset}
\label{app:mental_health_admissions_subset}
We operationalize mental-health-coded admissions using the International Classification of Diseases, Ninth Revision, Clinical Modification (ICD-9-CM): an admission qualifies if it contains at least one diagnosis code from Chapter 5, Mental Disorders (290–319) (Centers for Medicare \& Medicaid Services and National Center for Health Statistics, 2011) \cite{cms_icd9cm_2011}. Table~\ref{tab:icd9_mental_disorders} breaks down the key categories.

Our framework uses only free-text discharge summaries from the MIMIC-III
\texttt{NOTEEVENTS} table, which describe each patient's hospital stay.
We link each summary to structured MIMIC-III fields only to define the
study cohort and provide date anchors: diagnosis codes from
\texttt{DIAGNOSES\_ICD}, date of birth from \texttt{PATIENTS}, and
admission and discharge dates from \texttt{ADMISSIONS}. These structured
fields are not used as model inputs or training labels. The discharge
summaries are provided to the framework in their original form, without
chunking, section detection, or text normalization.


\begin{table}[H]
    \centering
    \small
    \begin{tabular}{
        @{}
        >{\raggedright\arraybackslash}p{0.20\columnwidth}
        >{\raggedright\arraybackslash}p{0.72\columnwidth}
        @{}
    }
        \toprule
        \textbf{Code range} & \textbf{Diagnostic category} \\
        \midrule
        \textbf{290--294} & Organic psychotic conditions \\
        \textbf{295--299} & Other psychoses \\
        \textbf{300--316} & Neurotic disorders, personality disorders, and other nonpsychotic mental disorders \\
        \textbf{317--319} & Intellectual disabilities \\
        \bottomrule
    \end{tabular}

    \vspace{2pt}

    \caption{Subcategories within ICD-9-CM Chapter 5, Mental Disorders
    (codes 290--319) \cite{cms_icd9cm_2011}. An admission is included in the mental-health-coded cohort if it contains at least one ICD-9-CM diagnosis code within the Mental Disorders chapter (290--319)}
    \label{tab:icd9_mental_disorders}
\end{table}
\FloatBarrier

\subsection{Benchmark Input Text Profile}
\label{app:benchmark_input_text_profile}
Because both model evaluation and human validation require comparison against the complete source discharge summary, input length determines the amount of narrative context that must be processed and reviewed.
Table~\ref{tab:benchmark_input_profile} summarizes the raw-text lengths of the 52-document benchmark. The summaries contain $1,774$ words on average (median: $1,624$; SD: $821$), ranging from $437$ to $3,583$ words and
totaling $92,222$ words. Character counts show similar variation, with a mean of $11,871$ characters and a range of $2,952$ to $24,808$. These statistics reflect the original discharge summaries supplied to the framework without chunking, section parsing, or text normalization, and illustrate the long-context processing and manual-review burden of the benchmark.

\begin{table}[t]
  \centering
  \scriptsize
  \setlength{\tabcolsep}{3.2pt}
  \renewcommand{\arraystretch}{1.12}
  \begin{tabular}{lrrrrrr}
    \toprule
    \textbf{Metric}
      & \textbf{Mean}
      & \textbf{Median}
      & \textbf{SD}
      & \textbf{Min}
      & \textbf{Max}
      & \textbf{Total} \\
    \midrule
    Characters
      & 11,871
      & 11,042
      & 5,481
      & 2,952
      & 24,808
      & 617,267 \\
    Words
      & 1,774
      & 1,624
      & 821
      & 437
      & 3,583
      & 92,222 \\
    \bottomrule
  \end{tabular}
  \caption{Raw-text profile of the 52 discharge summaries in the benchmark.}
  \label{tab:benchmark_input_profile}
\end{table}

\subsection{Token Usage for Silver-Standard Generation}
\label{app:token_usage_silver}
To quantify the inference required to construct the silver-standard supervision used in Section~\ref{downstream_eval_open_weight_models}, we recorded the API token-usage metadata returned by Gemini 2.5 Pro for the 1,000 admissions processed through the frozen three-stage framework. Across Stages 1 to 3, generation processed $45.13$M input tokens and produced $16.70$M output tokens, for a combined total of $61.84$M tokens, or $61,836$ tokens per admission on average. Stage 1 accounted for $25.66$M tokens ($41.5$\% of the total), Stage 2 for $19.52$M ($31.6$\%), and Stage 3 for $16.66$M ($26.9$\%). Table~\ref{tab:silver_token_usage} provides the complete stage-level breakdown.

These counts quantify the inference used to generate the $1,000$-example silver corpus in Section~\ref{downstream_eval_open_weight_models} and illustrate the computational demands of processing complete discharge summaries and their intermediate representations through a multi-stage long-context pipeline. The resulting corpus, however, can be reused across three downstream tasks and multiple open-weight student models, amortizing the initial teacher-generation cost. Of the reported input tokens, $16,177,254$ were served from cached context; because cached tokens are a subset of input tokens, they are not added again to the total.

\begin{table*}[t]
  \centering
  \small
  \setlength{\tabcolsep}{10pt}
  \renewcommand{\arraystretch}{1.3}

  \begin{tabular}{@{}lrrr@{}}
    \toprule
    \textbf{Stage}
      & \textbf{Input}
      & \textbf{Output}
      & \textbf{Total} \\
    \midrule

    Stage 1: Atomic Clinical Event Extraction
      & 20,464,371
      & 5,198,803
      & 25,663,174 \\

    Stage 2: Date \& Certainty Tagging
      & 9,794,222
      & 9,723,468
      & 19,517,690 \\

    Stage 3: Temporally Guided Summarization
      & 14,874,581
      & 1,780,523
      & 16,655,104 \\

    \midrule
    \textbf{Total}
      & \textbf{45,133,174}
      & \textbf{16,702,794}
      & \textbf{61,835,968} \\
    \bottomrule
  \end{tabular}

  \caption{Token usage for generating the 1,000-example silver-standard corpus with the frozen Gemini~2.5~Pro. Total tokens are the sum of input and output tokens. The grand total corresponds to an average of 61,836 tokens per admission across all three stages. Of the input tokens, 16,177,254 were served from cached context and are already included in the input counts.}
  \label{tab:silver_token_usage}
\end{table*}

\section{Agent Candidate Selection}
\label{app:agent_candidate_selection}
Following the selection of agents in Section~\ref{sec:implementation_setup}, a candidate is eligible if it supports a native context window of at least $32$K tokens which is sufficient to fit our P$99$ input without truncation. All candidates are evaluated with identical prompts and greedy decoding ($t=0$). 
An open-weight candidate additionally has to be (i) downloadable under a license permitting research and on-premise inference and (ii) loadable on our deployment stack across four L40S GPUs at bfloat16 precision. A proprietary candidate additionally has to be (iii) served through a managed endpoint compatible with our institutional data-use agreement (e.g., Amazon Bedrock or Google Vertex AI) and (iv) cost-feasible at corpus scale. A complete table of agent specifications can be found in Table~\ref{tab:model_agent_pool}.

Prior work recommends low decoding temperatures for classification and labelling tasks to stabilize outputs~\cite{Jin2024_WWW}. Because our stages demand both high precision and high recall, including sharp inclusion/exclusion at extraction and unambiguous tagging decisions downstream, we set temperature $t = 0$ (greedy decoding) for all three agentic stages. All other generation settings were held constant across models, and for open-weight models this produces byte-identical outputs across runs.


\begin{table*}[t]
  \centering
  \small
  \setlength{\tabcolsep}{5pt}
  \renewcommand{\arraystretch}{1.15}

  \begin{tabular}{
    @{}
    p{0.4\textwidth}
    p{0.2\textwidth}
    p{0.12\textwidth}
    p{0.1\textwidth}
    @{}
  }
    \toprule
    \textbf{Agent Model}
    & \textbf{Provider}
    & \textbf{\shortstack[l]{Params\\(total/active)}}
    & \textbf{Arch.} \\
    \midrule

    Gemini 2.5 Pro \cite{comanici2025gemini25pushingfrontier}
    & Google (via Vertex AI)
    & Undisclosed
    & n/r \\

    Gemini 2.5 Flash
    & Google (via Vertex AI)
    & Undisclosed
    & n/r \\

    Claude Sonnet 4.6 \cite{anthropic2026sonnet46}
    & Anthropic (via Bedrock)
    & Undisclosed
    & n/r \\

    Claude Haiku 4.5 \cite{anthropic2025haiku45}
    & Anthropic (via Bedrock)
    & Undisclosed
    & n/r \\

    Llama 3.3 70B \cite{meta2024llama33}
    & Meta
    & 70B
    & Dense \\

    Qwen 3.6 35B-A3B \cite{qwen2026qwen36}
    & Alibaba
    & 35B / 3B
    & MoE \\

    Gemma 4 26B-A4B \cite{gemmateam2026gemma4}
    & Google
    & 26B / 4B
    & MoE \\

    MedGemma 27B 
    \cite{sellergren2026medgemmatechnicalreport}
    & Google
    & 27B
    & Dense \\

    Mistral Small 3.2 24B \cite{mistral2025small32}
    & Mistral AI
    & 24B
    & Dense \\

    Meditron 3 (Qwen 2.5 7B) \cite{openmeditron2025meditron3}
    & OpenMeditron/EPFL
    & 7B
    & Dense \\

    Phi-4 Mini 3.8B \cite{abouelenin2025phi4mini}s
    & Microsoft
    & 3.B
    & Dense \\

    \bottomrule
  \end{tabular}
  \vspace{2pt}
  \caption{Model agents included in the candidate pool.
  Models vary in provider, domain specialization, serving environment,
  parameter count, and architecture.}
  \label{tab:model_agent_pool}
\end{table*}


\section{Benchmark Size Selection Via Power Analysis}
\label{app:power_analysis}
We fix the number of manually evaluated documents using an a priori power analysis for a chi-square test \cite{cohen_statistical_1988}. The required sample size is

\begin{equation}
    N = \frac{\lambda^{*}}{w^{2}},
\end{equation}

where \(w\) is Cohen's effect-size index and \(\lambda^{*}\) is the
noncentrality parameter at which the noncentral chi-square distribution
\(\chi^{2}(df,\lambda)\) reaches the target power relative to the
critical value \(\chi^{2}_{1-\alpha}(df)\).

For a pairwise comparison with one degree of freedom, we targeted a
large effect size (\(w = 0.5\)), a significance level of
\(\alpha = 0.05\), and power of \(1-\beta = 0.95\). This gives

\begin{equation}
    \lambda^{*}
    =
    \left(
        z_{1-\alpha/2} + z_{1-\beta}
    \right)^{2}
    =
    12.99,
\end{equation}

and therefore

\begin{equation}
    N
    =
    \frac{12.99}{0.5^{2}}
    =
    \frac{12.99}{0.25}
    =
    51.98.
\end{equation}

We round this value up to \(N = 52\) documents. Applying a
finite-population correction for the full corpus of 14,882 documents
produced an adjusted estimate of \(N_{\mathrm{adj}} = 51.80\), leaving
the required sample size effectively unchanged. At \(N = 52\), the
achieved power is \(0.950\).

The analysis is conducted using the \texttt{pwr} package
\cite{champely2020pwr} and independently verified using the noncentral chi-square distribution implemented in \texttt{SciPy} \cite{2020SciPy-NMeth}. The sampling unit is the discharge-summary document. Thus, the independence assumption applies across sampled documents, and correlations among events within a document do not artificially increase the effective sample size.

\section{Stage 1 \& Stage 2 Generation Prompts}
We present the concrete definitions for each of the sub-dimensions of timeline-robustness.
\label{app:stage1_stage2_gen_prompts}

We provide the complete prompts used for Stages 1 and 2 in
Tables~\ref{tab:appendix_stage1_prompt} and~\ref{tab:appendix_stage2_prompt}. Each stage
uses a fixed system prompt and a document-specific user-prompt template.
The system prompts were applied unchanged across all candidate models.
At inference time, only the bracketed placeholders were replaced with
the corresponding discharge summary, extracted events, and admission
anchors. No model-specific prompt tuning or manual preprocessing was
performed.

\subsection{Stage 1: Atomic Clinical Event Extraction}

The Stage 1 prompt instructs the model to extract all clinical
information from the discharge summary as atomic, self-contained
events. The prompt prioritizes coverage and source faithfulness:
repeated mentions are retained, negation is preserved, and the model is
instructed not to correct the note, infer unstated information, or add
clinical interpretation. Events are returned in their original document
order as an unnumbered list. The complete prompt is shown in
Table~\ref{tab:appendix_stage1_prompt}.

\subsection{Stage 2: ISO Date and Certainty Tagging}

The Stage 2 prompt assigns one temporal label to each event extracted in Stage 1. The model receives the full discharge summary as global context, together with the admission date, discharge date, date of birth, patient age, and the ordered list of atomic events. It is instructed to preserve the event text and assign exactly one of four labels:
\texttt{[EXACT]}, \texttt{[APPROX]}, \texttt{[PRE ADM]}, or
\texttt{[INDETERMINATE]}. The prompt follows a conservative decision rule: explicitly written dates receive \texttt{[EXACT]}, dates resolved from anchors or relative expressions receive \texttt{[APPROX]}, events framed as prior history receive \texttt{[PRE ADM]}, and
\texttt{[INDETERMINATE]} is used only when no defensible temporal resolution is available. The model is explicitly prohibited from guessing dates. The complete prompt is shown in
Table~\ref{tab:appendix_stage2_prompt}.

\begin{table*}[!htbp]
    \centering
    \small
    \begin{tabular}{p{0.10\linewidth} | p{0.85\linewidth}}
    \toprule
    \textbf{Type} & \textbf{Prompt}\\
    \toprule
         \textbf{System Prompt}&You are an atomic clinical event extraction engine. Your task is to extract a list of atomic clinical events (Presented as ATOMIC\_CLINICAL\_EVENTS) from an unstructured discharge summary.\\
         &\\
         &CLINICAL EVENTS: A clinical event is one clinically meaningful, patient-related unit of information documented in the note, expressed as a self-contained statement describing an action, condition, observation, or change in the patient’s clinical course.\\
         &\\
         &An event may describe something that: \\
         &1. Happened\\
         &2. Was observed\\
         &3. Was reported\\
         &4. Was performed\\
         &5. Was denied \\
         &6. Was known about the patient\\
         &\\
         &This includes, but is not limited to: Diagnoses, symptoms and signs, Procedures, lab results, imaging findings, medications, social history, behavioral events, family history, clinical decisions and plans, negated findings (these must be preserved exactly).\\
         &\\
         &ATOMIC CLINICAL EVENTS: We define an atomic clinical event as the smallest unit of clinical information expressible as a single subject-verb-object clause that cannot be further decomposed without loss of clinical meaning.\\
         &\\
         &Instructions:\\
         &1. Expand all abbreviations but keep the original in parentheses (e.g., ``shortness of breath (SOB)”, ``heart rate (HR) of 96”).\\
         &2. Extract every mention of every fact, even if the same fact appears in multiple sections. Do not deduplicate.\\
         &3. Do not correct errors, infer unstated information, or add clinical interpretation. Extract only what is explicitly written.\\
         &4. Negated findings must be preserved exactly (e.g., ``The patient has no thyromegaly").\\
         &5. Preserve the tense from the original text. For structured data without verbs (such as labs and vitals), use ``was.” \\
         &6. Output the atomic clinical events in document order as a list, with each item starting with ``- ". Do not number the bullets. Do not add explanations. Do not add any extra text.\\
         &\\
         \midrule
         \textbf{User Prompt}&DISCHARGE\_SUMMARY: <discharge\_summary>\\
         \bottomrule
    \end{tabular}
    \caption{Prompt used for atomic clinical event extraction}
    \label{tab:appendix_stage1_prompt}
\end{table*}

\newpage
\begin{table*}[!htbp]
    \centering
    \small
    \begin{tabular}{p{0.10\linewidth} | p{0.85\linewidth}}
    \toprule
    \textbf{Type} & \textbf{Prompt}\\
    \toprule
         \textbf{System Prompt}&You are a temporal tagger for atomic clinical events. Assign exactly one temporal tag to each event in the given ATOMIC\_CLINICAL\_EVENTS list, producing the TAGGED\_ATOMIC\_CLINICAL\_EVENTS list. Accurate tags let us reconstruct a patient's timeline from an unstructured note, so a wrong date is worse than a coarse one — NEVER invent or guess a date.\\
         &\\
         &You are given the original note (DISCHARGE\_SUMMARY) as global context and ANCHOR\_DATES (admission date, discharge date, and patient age if stated).\\
         &\\
         &DATE FORMATS: \\
         & \hspace{0.5em}- When a date is assigned, use the most specific format supported by the source: [YYYY-MM-DD], [YYYY-MM], or [YYYY]. \\
         & \hspace{0.5em}- For events spanning a range, use [START to END], where START and END can each be in any of the above formats (e.g., [2000-01-01 to 2000-01-02], [1998-11 to 2001-08-02]).\\
         & \hspace{0.5em}- Never add precision the source lacks: month-only resolves to [YYYY-MM]; year-only resolves to [YYYY].\\
         &\\
         &TAGS: assign the one option that applies.\\
         &[\textbf{EXACT}] -- a date (or range) is written in the event text itself and applies to it. Use it as written, normalized, regardless of where it falls: a dated pre-admission event (e.g. a past surgery) is EXACT, not [PRE ADM]; a future follow-up date is EXACT.\\
         & Examples include:\\
         & \hspace{0.5em}- “[2000-01-01] [EXACT] The patient underwent a surgical procedure on 2000-01-01”\\
         & \hspace{0.5em}- “[2010-03-04] [EXACT] On 2010-03-04 the patient was brought to the emergency department (ED)”\\
         & \hspace{0.5em}- “[1980-05-13] [EXACT] The patient is to follow up with his primary care provider on 1980-05-13” (a written date is EXACT even when it falls after discharge)\\
         &\\ [-6pt]
         &[\textbf{APPROX}] -- no explicit date is written in the event string, the event resolves to an anchor (admission, discharge, or DOB). Covers: the admission/presentation encounter; a relative offset from an anchor, at the offset's precision; an in-stay event with no specific date (use the [admission to discharge] range); a plan or instruction issued at discharge. Examples (admission 2010-01-06, discharge 2010-01-10):\\
         & Examples include:\\
         & \hspace{0.5em}- “[2010-01-01] [APPROX] 5 days before admission, the patient attempted suicide”\\
         & \hspace{0.5em}- “[2003] [APPROX] The patient quit smoking 7 years ago”\\
         & \hspace{0.5em}- “[2010-01-06 to 2010-01-10] [APPROX] A procedure was held during the hospitalization”\\
         & \hspace{0.5em}- “[2010-01-06] [APPROX] In the emergency department (ED), the patient did well” (a prior related event established the ED visit was on 2010-01-06; events tied to that visit resolve to it)\\
         &\\ [-6pt]
         &[\textbf{PRE ADM}] -- the event is framed as prior to or continuous through this admission — including but not necessarily limited to past medical history, “history of …”, “status post”/“s/p”, prior events, family or social history with no exact or relative time mentions, demographics, or an undated chronic state. Carries no [DATE] tag in front. \\
         & Examples include:\\
         & \hspace{0.5em}- “[PRE ADM] The patient has a history of depression”\\
         & \hspace{0.5em}- “[PRE ADM] The patient is a 37 year old female”\\
         & \hspace{0.5em}- “[PRE ADM] The patient was involved in an accident”\\
         & \hspace{0.5em}- “[PRE ADM] The patient’s mother died at age 40 of cancer”.\\
         &\\ [-6pt]
         & Strict Guideline\\
         &Prefer the three tags above. Use the last option, [INDETERMINATE], only as a last resort.\\
         &\\ [-6pt]
         &[\textbf{INDETERMINATE}] -- use ONLY after [EXACT], [APPROX], and [PRE ADM] have each been ruled out: no written date (not EXACT), no resolution to any anchor (not APPROX), and no indication it occurred before admission (not PRE ADM). The timing genuinely cannot be determined. This is not a fallback for uncertainty — if any of the first three plausibly applies, use that one. Carries no date. An example includes:\\
         & \hspace{0.5em}- “[INDETERMINATE] An outpatient ultrasound was recommended at some point”\\
         &\\
         &Output Instructions\\
         & \hspace{0.5em}1. Preserve the exact wording of each input event in the output. Do not reword, reorder, correct errors, infer unstated facts, or add interpretation.\\
         & \hspace{0.5em}2. One bullet per event, starting with “- ”. Tag block is “[DATE] [TAG]” for EXACT and APPROX, and the tag alone (“[PRE ADM]” or “[INDETERMINATE]”) for the two that carry no date.\\
         & \hspace{0.5em}3. Do not number bullets. Do not add any text outside the bullets.\\
         \midrule
         \textbf{User Prompt}
         &ANCHOR\_DATES:\\[-8pt]
         & \hspace{0.5em}- Date of Birth: {date\_of\_birth}\\
         & \hspace{0.5em}- Admission Date: <admission\_date>\\
         & \hspace{0.5em}- Discharge Date: <discharge\_date>\\
         &\\ [-8pt]
         &DISCHARGE\_SUMMARY: <discharge\_summary>\\
         &ATOMIC\_CLINICAL\_EVENTS: <atomic\_clinical\_events>\\
         \bottomrule
    \end{tabular}
    \caption{Prompt used for date \& certainty tagging}
    \label{tab:appendix_stage2_prompt}
\end{table*}
\FloatBarrier

\clearpage
\section{Stage 1 \& Stage 2 Evaluation Rubrics}
\label{app:stage1_stage2_eval_rubrics}
Stage 2 is evaluated using four complementary dimensions. Completeness measures whether the extracted timeline captures the source-grounded clinical events in the discharge summary. Semantic faithfulness assesses whether each event preserves the meaning of the source, including negation, uncertainty, quantities, and other qualifiers. Temporal-tag accuracy evaluates both the assigned certainty category and the resolved date or date range. Anti-hallucination identifies events that introduce claims unsupported by the source note. It's important to note that anti-hallucination was evaluated only by human annotators. Unlike the other dimensions, this criterion requires determining whether any claim in an extracted event lacks support anywhere in the source note. Prior work shows that LLM-as-judge reliability varies substantially across tasks and evaluation properties, and that even strong LLMs remain imperfect at recognizing hallucinated content \cite{bavaresco-etal-2025-llms, li-etal-2023-halueval}. This limitation is especially consequential in the clinical domain, where subtle, plausible unsupported claims may be difficult to distinguish from source-grounded information. Because a missed hallucination would directly contaminate the clinician-verified reference set, we reserved this criterion for exhaustive human review rather than treating an LLM's factuality judgment as ground truth.

All rubrics are included in Tables~\ref{tab:appendix_stage2_completeness} to~\ref{tab:appendix_stage2_hallucination}.
\begin{table*}[!htbp]
    \centering
    \small
    \setlength{\tabcolsep}{5pt}
    \renewcommand{\arraystretch}{1.08}

    \begin{tabular}{p{0.16\linewidth} | p{0.79\linewidth}}
        \toprule
        \textbf{Component} & \textbf{Dimension A: Completeness} \\
        \toprule

        \textbf{Guiding Question}
        &
        Did the extraction capture the source-grounded events from the note?
        \\

        &\\[-4pt]

        \textbf{Definition}
        &
        A source-grounded clinical event is a single, self-contained assertion
        documented in the note and concerning the patient's care. This includes,
        but is not limited to, conditions, findings, procedures, medications,
        laboratory and vital-sign values, plans and instructions, demographics,
        allergies, admission and discharge facts, and family or social history.
        \\

        &\\[-4pt]

        &
        An assertion is source-grounded when it is stated in the note; it does
        not need to be clinically salient. Clinical salience is not assessed
        under this rubric.
        \\

        &\\[-4pt]

        \textbf{Exclusions}
        &
        An assertion is excluded only when its entire content is administrative,
        such as a clinician or provider name alone, telephone or fax number,
        pager number, address, ward or room identifier, or document-formatting
        artifact. An administrative detail appearing within an otherwise
        clinical assertion does not exclude that assertion.
        \\

        &\\[-4pt]

        \textbf{Reporting}
        &
        Report every missing event, including its source span in the note and a
        one-line description. The expected-event set is the union of the
        source-grounded candidate events and the missing events reported by the
        evaluator. Coverage is calculated as the number of captured expected
        events divided by the size of this union.
        \\

        \midrule

        \textbf{Score 5}
        &
        The candidate list captures at least 95\% of the expected
        source-grounded clinical events.
        \\

        \textbf{Score 4}
        &
        The candidate list captures at least 80\% but less than 95\% of the
        expected source-grounded clinical events.
        \\

        \textbf{Score 3}
        &
        The candidate list captures at least 60\% but less than 80\% of the
        expected source-grounded clinical events.
        \\

        \textbf{Score 2}
        &
        The candidate list captures at least 40\% but less than 60\% of the
        expected source-grounded clinical events.
        \\

        \textbf{Score 1}
        &
        The candidate list captures less than 40\% of the expected
        source-grounded clinical events.
        \\

        \bottomrule
    \end{tabular}

    \caption{Stage-2 evaluation rubric for Dimension A: Completeness.}
    \label{tab:appendix_stage2_completeness}
\end{table*}

\begin{table*}[!htbp]
    \centering
    \small
    \setlength{\tabcolsep}{5pt}
    \renewcommand{\arraystretch}{1.08}

    \begin{tabular}{p{0.16\linewidth} | p{0.79\linewidth}}
        \toprule
        \textbf{Component} & \textbf{Dimension B: Semantic Faithfulness} \\
        \toprule

        \textbf{Guiding Question}
        &
        Does each candidate event preserve the clinical meaning of the source
        note while allowing the permitted normalizations?
        \\

        &\\[-4pt]

        \textbf{Scope}
        &
        Judge only the event text. Ignore the bracketed date and temporal-tag
        block preceding it, because temporal tags are evaluated separately.
        Within this dimension, assess only whether temporal expressions written
        inside the event text match those written in the source note.
        \\

        &\\[-4pt]

        \textbf{Definition}
        &
        An event preserves the source meaning only when all of the following
        properties survive rewriting: polarity or negation; epistemic status,
        including hedging and expressions such as ``likely'' or ``concern for'';
        experiencer, such as the patient versus a family member; quantities,
        units, doses, and laterality; severity, chronicity, and frequency;
        temporal expressions appearing in the text; and conditionality or
        hypotheticality.
        \\

        &\\[-4pt]

        &
        Each event receives a binary semantic judgment. A wording change that
        preserves all meaning-invariant properties is not an error.
        \\

        &\\[-4pt]

        \textbf{Hallucinations}
        &
        Events with no identifiable source assertion are hallucinations and are
        evaluated under Dimension D. They are excluded from the denominator for
        Dimension B. No event may be counted as an error under both Dimensions B
        and D.
        \\

        &\\[-4pt]

        \textbf{Reporting}
        &
        Report every event that fails, including its event ID, corresponding
        source span, and one error code from the following list:
        \\

        &
        \textbullet\ \texttt{negation\_lost}: a negated finding is asserted,
        or an asserted finding is negated.
        \\

        &
        \textbullet\ \texttt{uncertainty\_lost}: a hedged claim is asserted
        as fact, or a factual claim is incorrectly hedged.
        \\

        &
        \textbullet\ \texttt{experiencer\_shifted}: a statement concerning a
        family member or another person is attributed to the patient, or vice
        versa.
        \\

        &
        \textbullet\ \texttt{value\_or\_unit\_altered}: a quantity, unit,
        dose, count, or laterality differs from the source.
        \\

        &
        \textbullet\ \texttt{qualifier\_altered}: severity, chronicity, or
        frequency differs from the source.
        \\

        &
        \textbullet\ \texttt{temporal\_cue\_altered}: a temporal expression
        in the event text differs from or is omitted relative to the source.
        \\

        &
        \textbullet\ \texttt{conditionality\_lost}: a conditional or
        hypothetical statement is presented as an actual occurrence.
        \\

        &
        \textbullet\ \texttt{abbreviation\_misexpanded}: an abbreviation is
        expanded to a meaning unsupported by the source.
        \\

        &
        \textbullet\ \texttt{other\_distortion}: the meaning differs for a
        reason not covered by the other error codes.
        \\

        &\\[-4pt]

        \textbf{Normalization}
        &
        Separately record events whose wording deviates from the permitted
        normalizations while preserving all meaning-invariant properties using
        the flag \texttt{normalization\_deviation}. These cases are not semantic
        errors and do not affect the score.
        \\

        \midrule

        \textbf{Score 5}
        &
        At least 95\% of candidate events preserve the source meaning.
        \\

        \textbf{Score 4}
        &
        At least 80\% but less than 95\% of candidate events preserve the
        source meaning.
        \\

        \textbf{Score 3}
        &
        At least 60\% but less than 80\% of candidate events preserve the
        source meaning.
        \\

        \textbf{Score 2}
        &
        At least 40\% but less than 60\% of candidate events preserve the
        source meaning.
        \\

        \textbf{Score 1}
        &
        Less than 40\% of candidate events preserve the source meaning.
        \\

        \bottomrule
    \end{tabular}

    \caption{Stage-2 evaluation rubric for Dimension B: Semantic Faithfulness.}
    \label{tab:appendix_stage2_semantic}
\end{table*}

\begin{table*}[!htbp]
    \centering
    \small
    \setlength{\tabcolsep}{5pt}
    \renewcommand{\arraystretch}{1.08}

    \begin{tabular}{p{0.16\linewidth} | p{0.79\linewidth}}
        \toprule
        \textbf{Component} & \textbf{Dimension C: Temporal-Tag Accuracy} \\
        \toprule

        \textbf{Guiding Question}
        &
        For each candidate event, is the assigned temporal tag correct, and is
        the assigned date correct?
        \\

        &\\[-4pt]

        \textbf{Tag Format}
        &
        Every event carries exactly one temporal tag. \texttt{[EXACT]} and
        \texttt{[APPROX]} also carry a date or date range.
        \texttt{[PRE ADM]} and \texttt{[INDETERMINATE]} carry no date.
        \\

        &\\[-4pt]

        \textbf{\texttt{[EXACT]}}
        &
        A date is written in the event text itself. A date whose month and day
        are written and whose year is resolved from the note is also
        \texttt{[EXACT]}. For example, ``on 5-9'' in a note with admission year
        2135 becomes \texttt{[2135-05-09] [EXACT]}. An event whose text contains
        no date is never \texttt{[EXACT]}.
        \\

        &\\[-4pt]

        \textbf{\texttt{[APPROX]}}
        &
        No date is written in the event text, but the event resolves to the
        admission date, discharge date, date of birth, a relative offset from
        one of these anchors, or a date established elsewhere in the note. An
        in-stay event without a specific date receives the admission-to-discharge
        range. A plan issued at discharge receives the discharge date.
        \\

        &\\[-4pt]

        \textbf{\texttt{[PRE ADM]}}
        &
        The event is framed as occurring before or continuing through the
        current admission, with no written or resolvable date. This includes
        history, ``status post'' statements, demographics, undated chronic
        states, and undated family or social history.
        \\

        &\\[-4pt]

        \textbf{\texttt{[INDETERMINATE]}}
        &
        None of the other categories applies and the event's timing cannot be
        determined. This is a last-resort label.
        \\

        &\\[-4pt]

        \textbf{Date Precision}
        &
        A date is correct only when it matches the date that should have been
        assigned at the same level of precision. Assigning
        \texttt{[2135-05]} when \texttt{[2135-05-09]} was recoverable is an
        error, as is assigning \texttt{[2135-05-09]} when only
        \texttt{[2135-05]} was supported. For a date range, both endpoints must
        match.
        \\

        \midrule

        \textbf{Event Score 1.0}
        &
        Assign 1.0 when the temporal tag is correct and, for
        \texttt{[EXACT]} and \texttt{[APPROX]}, the date is also correct.
        \texttt{[PRE ADM]} and \texttt{[INDETERMINATE]} receive 1.0 when the
        tag is correct because they carry no date.
        \\

        \textbf{Event Score 0.5}
        &
        Assign 0.5 only when both the assigned and correct tags are
        \texttt{[EXACT]} or \texttt{[APPROX]}, and exactly one of the tag or
        date is correct. This includes cases where the tag is correct but the
        date is wrong, or the date is correct but \texttt{[EXACT]} and
        \texttt{[APPROX]} are interchanged.
        \\

        \textbf{Event Score 0.0}
        &
        Assign 0.0 in every other case. This includes assigning a no-date tag
        when a date-bearing tag is correct; assigning a date-bearing tag when a
        no-date tag is correct; confusing \texttt{[PRE ADM]} with
        \texttt{[INDETERMINATE]}; assigning both the wrong tag and wrong date;
        or emitting \texttt{[EXACT]} or \texttt{[APPROX]} without a date.
        \\

        \bottomrule
    \end{tabular}

    \caption{Stage-2 evaluation rubric for Dimension C: Temporal-Tag
    Accuracy (Part 1 of 2).}
    \label{tab:appendix_stage2_temporal}
\end{table*}

\begin{table*}[!htbp]
    \ContinuedFloat
    \centering
    \small
    \setlength{\tabcolsep}{5pt}
    \renewcommand{\arraystretch}{1.08}

    \begin{tabular}{p{0.16\linewidth} | p{0.79\linewidth}}
        \toprule
        \textbf{Component} & \textbf{Dimension C: Temporal-Tag Accuracy} \\
        \toprule

        \textbf{Reporting}
        &
        Report every event receiving a score below 1.0. Include its event ID,
        the temporal tag and date believed to be correct, and one error code
        from the following list:
        \\


        &
        \textbullet\ \texttt{tag\_over\_specific}: \texttt{[EXACT]} or
        \texttt{[APPROX]} was assigned where \texttt{[PRE ADM]} or
        \texttt{[INDETERMINATE]} was correct, meaning that a date was invented.
        \\

        &
        \textbullet\ \texttt{tag\_under\_specific}: \texttt{[PRE ADM]} or
        \texttt{[INDETERMINATE]} was assigned where a date was recoverable.
        \\

        &
        \textbullet\ \texttt{tag\_exact\_approx\_confused}:
        \texttt{[EXACT]} and \texttt{[APPROX]} were interchanged.
        \\

        &
        \textbullet\ \texttt{tag\_preadm\_indet\_confused}:
        \texttt{[PRE ADM]} and \texttt{[INDETERMINATE]} were interchanged.
        \\

        &
        \textbullet\ \texttt{date\_wrong\_value}: the assigned date contains
        the wrong day, month, or year.
        \\

        &
        \textbullet\ \texttt{date\_too\_coarse}: the date is correct but less
        precise than the note supports.
        \\

        &
        \textbullet\ \texttt{date\_too\_precise}: the date is more precise
        than the note supports.
        \\

        &
        \textbullet\ \texttt{date\_range\_wrong}: one or both endpoints of a
        date range are incorrect.
        \\

        &
        \textbullet\ \texttt{date\_missing}: \texttt{[EXACT]} or
        \texttt{[APPROX]} carries no date.
        \\

        &
        \textbullet\ \texttt{date\_wrongly\_present}: \texttt{[PRE ADM]} or
        \texttt{[INDETERMINATE]} incorrectly carries a date.
        \\

        &
        \textbullet\ \texttt{other\_temporal\_inaccuracy}: another temporal
        error, accompanied by an explanation.
        \\

        &\\[-4pt]

        \textbf{Aggregate Rate}
        &
        Temporal-tag accuracy is calculated as the sum of the per-event scores
        divided by the number of assessed events, expressed as a percentage.
        \\

        \midrule

        \textbf{Score 5}
        &
        At least 95\% temporal-tag accuracy.
        \\

        \textbf{Score 4}
        &
        At least 80\% but less than 95\% temporal-tag accuracy.
        \\

        \textbf{Score 3}
        &
        At least 60\% but less than 80\% temporal-tag accuracy.
        \\

        \textbf{Score 2}
        &
        At least 40\% but less than 60\% temporal-tag accuracy.
        \\

        \textbf{Score 1}
        &
        Less than 40\% temporal-tag accuracy.
        \\

        &\\[-4pt]

        \textbf{Rate Calculation}
        &
        For Dimensions A--C, calculate rates to one decimal place. Do not round
        the rate before assigning its score band. Report the counts used in the
        calculation.
        \\

        \bottomrule
    \end{tabular}

    \caption[]{Stage-2 evaluation rubric for Dimension C: Temporal-Tag
    Accuracy (Part 2 of 2).}
\end{table*}

\begin{table*}[!htbp]
    \centering
    \small
    \setlength{\tabcolsep}{5pt}
    \renewcommand{\arraystretch}{1.08}

    \begin{tabular}{p{0.16\linewidth} | p{0.79\linewidth}}
        \toprule
        \textbf{Component} & \textbf{Dimension D: Anti-Hallucination} \\
        \toprule

        \textbf{Guiding Question}
        &
        Does every factual claim in the candidate events come from the source
        note?
        \\

        &\\[-4pt]

        \textbf{Definition}
        &
        This dimension detects information added by the extraction. It does not
        assess information that was changed from an existing source assertion,
        and it does not penalize permitted rewrites.
        \\

        &\\[-4pt]

        \textbf{Decision Rule}
        &
        For each candidate event, determine whether the source note contains an
        assertion of which the event is a version, even if the event distorts
        that assertion. If such an assertion exists, the event is not a
        hallucination and any meaning error belongs under Dimension B. If no
        corresponding source assertion exists, the event is a hallucination.
        \\

        &\\[-4pt]

        \textbf{Hallucination Types}
        &
        \textbullet\ \texttt{fabricated\_event}: an action, finding,
        procedure, or occurrence that the source note never states.
        \\

        &
        \textbullet\ \texttt{fabricated\_attribute}: a value, unit, dose,
        count, side, or qualifier attached to a real event when the source gives
        no such attribute. When the source gives a different value for a known
        event, the error belongs under Dimension B.
        \\

        &
        \textbullet\ \texttt{fabricated\_entity}: a medication, test,
        diagnosis, or person that the source note never names.
        \\

        &
        \textbullet\ \texttt{fabricated\_date}: a date written within the
        event text that does not appear in the source note. Errors in the
        bracketed date or temporal-tag block belong under Dimension C.
        \\

        &
        \textbullet\ \texttt{unwarranted\_inference}: a causal, diagnostic,
        or prognostic conclusion not explicitly stated in the source note.
        Clinical reasoning stated in the source may be extracted, but
        unsupported reasoning introduced by the model may not.
        \\

        &\\[-4pt]

        \textbf{Do Not Assess}
        &
        Do not assess temporal tags or bracketed dates, which belong under
        Dimension C; meaning changes to events that have an identifiable source,
        which belong under Dimension B; or any permitted rewrite.
        \\

        &\\[-4pt]

        \textbf{Reporting}
        &
        Report every hallucinated event, including its event ID and complete
        event text.
        \\

        \midrule

        \textbf{Score 1}
        &
        No candidate event contains a hallucination.
        \\

        \textbf{Score 0}
        &
        At least one candidate event contains a hallucination.
        \\

        \bottomrule
    \end{tabular}

    \caption{Stage-2 evaluation rubric for Dimension D:
    Anti-Hallucination.}
    \label{tab:appendix_stage2_hallucination}
\end{table*}

\section{Stage 1 \& Stage 2 Evaluation Platform}
\label{app:stage1_stage2_eval_platform}
\begin{figure*}[t]
    \centering
    \includegraphics[width=\textwidth]{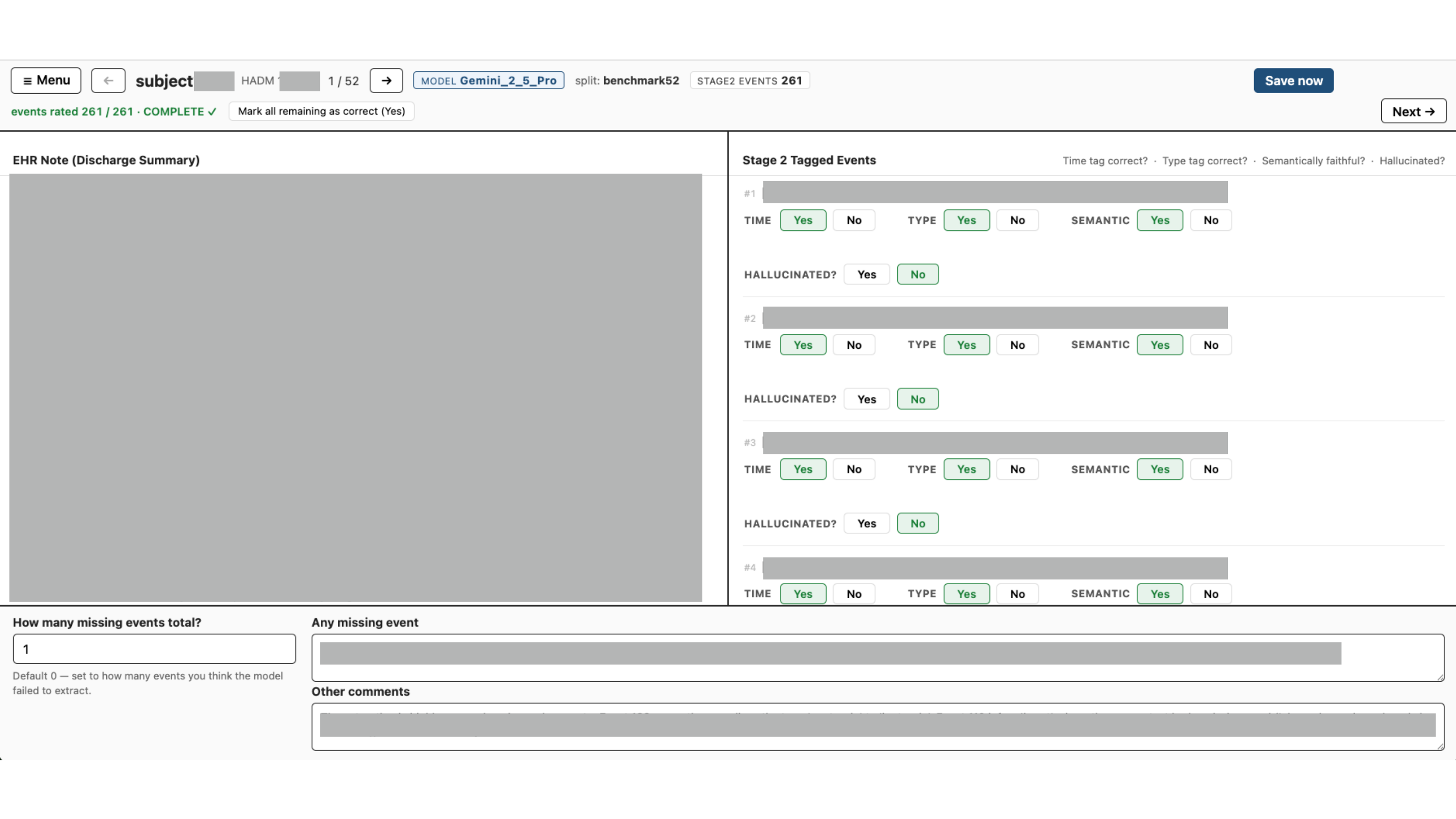}
    \caption{Web interface of the Stage-2 event-level human annotation platform. A single subject is shown: the source discharge summary (left) beside the model's Stage-2 tagged event list (right), with per-event Yes/No controls for the four rubric dimensions (time tag, type tag, semantic faithfulness, hallucination) and a footer for logging missed events. All protected health information and note/event text are redacted (gray bars) for this figure.}
    \label{fig:appendix_stage2_annotation_platform}
\end{figure*}

Human evaluation of Stage-2 outputs was conducted through a purpose-built local web application (Figure~\ref{fig:appendix_stage2_annotation_platform}). The tool runs as a self-contained Python program with a static HTML interface, requires no external services or database, and is accessed through SSH port forwarding so protected health information remains on the host machine. Model outputs are loaded into an ordered annotation queue, and annotators resume from their first unfinished document.

The interface displays the full discharge summary alongside the model-generated events in source order, including their temporal and certainty tags. For each event, annotators independently assess temporal accuracy, certainty-tag accuracy, semantic faithfulness, and hallucination, with criterion-specific comment fields for flagged errors. To capture omissions, the interface also records the number of missing events and the corresponding source passages.

The platform includes progress tracking, bulk confirmation for all-correct events, checks that prevent advancing with incomplete ratings, keyboard navigation, and immediate crash-safe saving to per-annotator JSON files. Completed work can therefore be interrupted and resumed without loss.

\textbf{\section{Stage 1 \& Stage 2 Evaluation Agreement and Error Analysis}
\label{sec:stage1_stage2_eval_agreement}}
Table~\ref{tab:stage2_judge_clinician_agreement} reports agreement between the
Gemini 3.1 Pro evaluator and clinician-guided human annotations across
15,891 Stage-2 events from 52 discharge summaries. We report Cohen's
\(\kappa\), Gwet's AC1, and positive and negative agreement. Because
confirmed errors are rare relative to correct events, Cohen's
\(\kappa\) is sensitive to class prevalence; Gwet's AC1 and the
class-specific agreement measures provide complementary views of
agreement under this imbalance.

Overall agreement is high on the dominant non-error class:
negative agreement reaches $0.986$, while Gwet's AC1 is
$0.972$. Agreement is lower on the rare error class, with
positive agreement of $0.669$. This indicates that the evaluator and human reviewers agree strongly on which events are correct, but differ more often when identifying or characterizing errors.

The disagreements are concentrated in temporal accuracy rather than
semantic faithfulness. Although human annotation and the automated
judge report nearly identical total numbers of temporal errors
($608$ versus $602$), an event-level breakdown reveals a strong directional asymmetry: human review identified 175 over-specified tags on pre-admission medications and demographics, 169 of which the judge
failed to flag, whereas judge-only findings were concentrated in
overlapping date-range and date-granularity codes (Appendix~\ref{app:stage2_temporal_error_types_analysis}). Temporal-accuracy agreement reaches \(\kappa=0.617\), whereas semantic-faithfulness negative agreement is \(1.000\). Manual review identified two recurring sources of temporal disagreement. First, some models anchor home medications to the admission date using \texttt{[APPROX]}, while clinicians classify them as \texttt{[PRE ADM]}; the automated evaluator does not consistently penalize this distinction. Second, the evaluator sometimes accepts a coarse date range even when the event can be resolved to a more specific in-stay anchor. Thus, disagreement is concentrated in borderline cases of pre-admission status and date granularity rather than being distributed uniformly across the task.

The integrity of the gold set does not depend on evaluator agreement.
Human annotators exhaustively reviewed all 15,891 events under clinician guidance and corrected every confirmed error after escalating errors to discussion with clinicians, regardless of whether the automated evaluator detected it. This process produced 52  human-corrected pairs of discharge summaries and tagged timelines.

\subsection{Temporal Error-Type Analysis}
\label{app:stage2_temporal_error_types_analysis}

To characterize the lower positive agreement on temporal accuracy, we compared the finalized event-level error codes assigned through
exhaustive human review with those assigned by the calibrated LLM
judge across the 52-note benchmark. 

\begin{figure*}[!t]
    \centering
    \includegraphics[width=\textwidth]
    {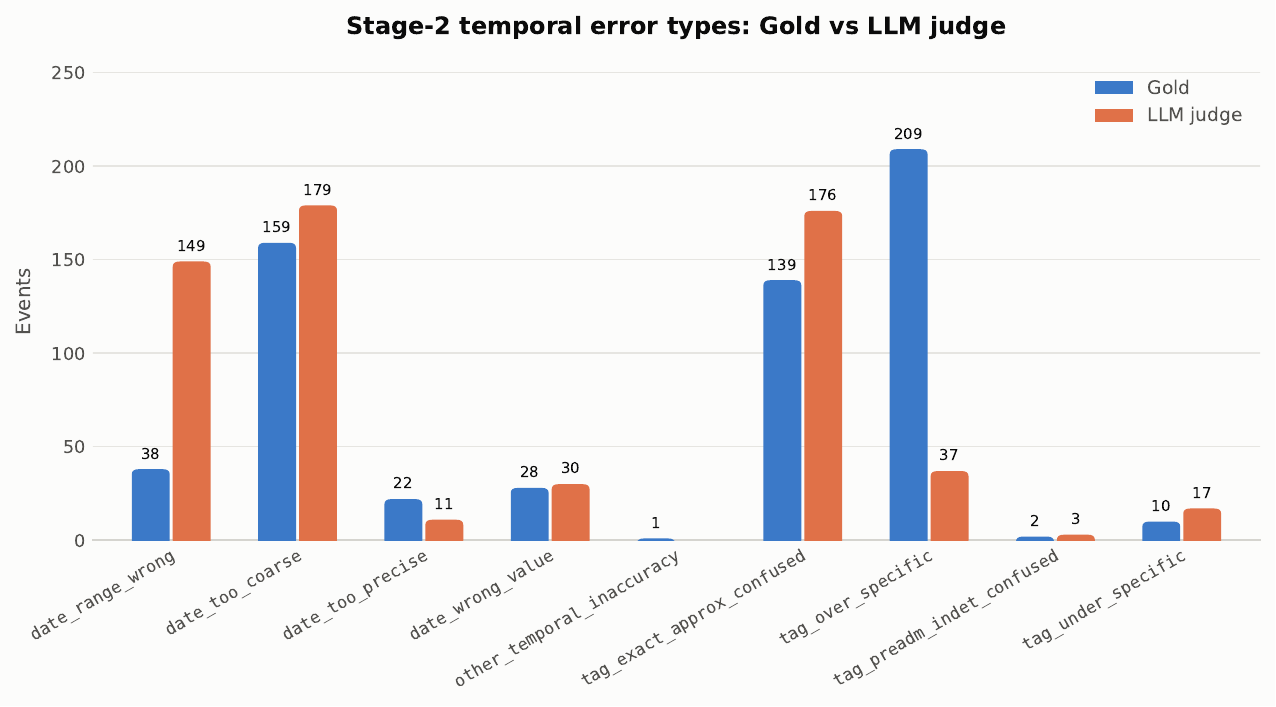}
    \caption{Stage-2 temporal-error counts from human annotation and
    the calibrated LLM judge. Despite similar totals (608 versus 602),
    human review surfaced 175 over-specified tags on pre-admission
    medications and demographics that the judge almost entirely missed,
    while judge-only errors concentrated in overlapping date-range and
    granularity categories.}
    \label{fig:stage2_temporal_error_types}
\end{figure*}

As shown in Figure~\ref{fig:stage2_temporal_error_types}, the aggregate counts are similar: human annotation identified 608 temporal errors, compared with 602 identified by the judge. Human review found at least one temporal error in 50 of the 52 notes, whereas the judge identified an error in 46. 

The clearest asymmetry concerns over-specified temporal tags. Human annotation identified 175 such errors involving pre-admission medications (151 events) and patient demographics (24 events), of which the judge failed to flag 169 (96.6\%). These are not cases in
which the human and judge applied different error codes to the same behavior: the judge returned no temporal error at all. The pattern therefore represents a systematic false-negative class in which the
judge accepts admission-anchored \texttt{[EXACT]} or
\texttt{[APPROX]} labels for standing or pre-admission facts that should receive \texttt{[PRE ADM]}.

The reverse asymmetry is qualitatively different. Of the 200 judge-only errors, 80 were coded as \texttt{date\_range\_wrong} and
56 as \texttt{date\_too\_coarse}. These codes describe closely related aspects of temporal granularity and may place the boundary between an incorrect range and an insufficiently precise date differently. Thus, similar aggregate error counts do not indicate equivalent error detection: human review exposes a clinically coherent
error class that the judge largely misses, whereas much of the judge's excess is concentrated around rubric boundaries for range and granularity.

\subsection{Other Error-Types and Analysis}
\label{app:stage2_other_error_types_analysis}
The $21$ non-temporal errors are  minor and mechanical, encompassing just $0.13\%$ of all $15,891$ events. Unlike the temporal errors, almost none reflect clinical misunderstanding. The dominant failure mode is over-precision: the model quietly hardens hedged or masked source text into confident assertions. Roughly half of the semantic errors convert uncertainty into fact (e.g., ``? <*clinical event*>" becomes ``possible <*clinical event*>"), or add unsupported specificity to medications (such as "twice a day", "twice daily" appended to a drug the note never dosed). The rest stem from misreading MIMIC's de-identification masks and OCR-garbled tokens. The 7 hallucinations are similar in spirit — mostly home/admission medications and lab values carrying detail not fully grounded in the source — and the single completeness miss (1 missing patient date of birth event across 52 notes) is negligible. In short, on everything except temporal tagging the model is essentially faithful; its only residual weakness is a tendency to resolve ambiguity in the source rather than preserve it.

\begin{table}[t]
  \centering
  \footnotesize
  \setlength{\tabcolsep}{3pt}
  \renewcommand{\arraystretch}{1.08}

  \begin{tabular}{@{}lcccc@{}}
    \toprule
    \textbf{Dimension}
    & \textbf{Cohen's $\kappa$}
    & \textbf{AC1}
    & \textbf{Pos.}
    & \textbf{Neg.} \\
    \midrule
    \textbf{Overall}
    & \textbf{0.655}
    & 0.972
    & 0.669
    & 0.986 \\

    Temporal accuracy
    & 0.617
    & 0.971
    & 0.631
    & 0.986 \\

    Semantic faithfulness
    & 0.611
    & 0.999
    & 0.611
    & 1.000 \\
    \bottomrule
  \end{tabular}

  \vspace{2pt}
  \caption{Agreement between the Gemini 3.1 Pro and gold annotation, calculated over 52 discharge summaries producing 15,891 events. AC1 denotes Gwet's AC1; Pos.\ and Neg.\ denote positive and negative agreement on the rare error and dominant non-error classes, respectively. Anti-hallucination is excluded because it was evaluated only by human annotators.}
  \label{tab:stage2_judge_clinician_agreement}
\end{table}

\section{Stage 3 Generation Prompt}
\label{app:stage3_gen_prompts}
The Stage 3 prompt converts the tagged event timeline into a chronological prose summary. It specifies how temporal tags should be rendered, which clinical information should be included or excluded, how duplicate or overlapping events should be combined, and how factual and temporal fidelity should be preserved. Because of its length, the complete prompt is presented across three consecutive sub-tables (Table~\ref{tab:appendix_stage3_prompt}).

\begin{table*}[!htbp]
    \centering
    \small
    \setlength{\tabcolsep}{5pt}
    \renewcommand{\arraystretch}{1.05}

    \begin{tabular}{p{0.10\linewidth} | p{0.85\linewidth}}
    \toprule
    \textbf{Type} & \textbf{Prompt}\\
    \toprule

    \textbf{System Prompt}
    &
    You will convert a tagged clinical event list into a temporally
    structured prose summary.\\

    &\\[-4pt]
    &\textbf{INPUTS}\\

    &You are given two inputs:\\

    &1. \textbf{ORIGINAL DISCHARGE SUMMARY} --- the source note.
    Use it only to clarify wording, expand abbreviations, or resolve
    unclear phrasing. Do not take any fact from it that is not present
    in the tagged event list.\\

    &2. \textbf{TAGGED EVENT LIST} --- the authoritative source.
    Every fact and every date in the summary must come from this list.\\

    &\\[-4pt]
    &\textbf{TEMPORAL RENDERING RULES
    (mandatory, no exceptions)}\\

    &Each event carries a temporal tag. Render each tag exactly as
    specified:\\

    &\textbullet\ \texttt{[YYYY-MM-DD] [EXACT]}:
    ``On YYYY-MM-DD, \ldots''\\

    &\textbullet\ \texttt{[YYYY-MM-DD] [APPROX]}:
    ``Approximately on YYYY-MM-DD, \ldots''\\

    &\textbullet\
    \texttt{[YYYY-MM-DD to YYYY-MM-DD] [APPROX]}:
    ``Approximately from YYYY-MM-DD to YYYY-MM-DD, \ldots''\\

    &\textbullet\ \texttt{[YYYY] [EXACT]}:
    ``In YYYY, \ldots''\\

    &\textbullet\ \texttt{[PRE ADM]}:
    ``Before admission, \ldots''\\

    &\\[-4pt]
    &Rules:\\

    &\textbullet\ Write all dates in ISO format
    (\texttt{YYYY-MM-DD}), zero-padded, exactly as given in the tag.\\

    &\textbullet\ The word ``approximately'' applies to the date,
    not to the event. The event is certain; only its timing is
    approximate. Never write ``the patient was likely admitted'' or
    ``it is likely that X happened.''\\

    &\textbullet\ Reproduce each event's negations and hedges exactly
    as written; do not add, delete, or replace them.\\

    &\textbullet\ If the event text restates timing already supplied
    by the tag, such as ``admitted on 2122-5-19'' or ``on postoperative
    day two,'' delete the repeated timing phrase from the sentence.\\

    &\textbullet\ You may adjust grammar and tense minimally so that
    the sentence reads correctly after the temporal prefix. Do not
    change the clinical content.\\

    &\\[-4pt]
    &\textbf{DEDUPLICATION AND MERGING}\\

    &\textbullet\ \textbf{Exact duplicates:} If two events have the
    same tag and the same content, output the content once.\\

    &\textbullet\ \textbf{Subsumption:} If one event's content is
    fully contained in another's, such as ``The patient is female''
    and ``The patient is a 19-year-old female,'' output only the more
    specific event.\\

    &\textbullet\ \textbf{Merging:} Events with identical tags
    (the same date and the same type) must be combined under one
    temporal prefix and joined into one or a few sentences. Never
    repeat the prefix, and never merge events across different dates
    or types. This includes recombining events split from one clinical
    statement, such as ``Her incision was clean,'' ``was dry,'' and
    ``was intact.''\\

    &\textbullet\ Merge date of birth, age, and sex into one opening
    demographic sentence.\\

    \bottomrule
    \end{tabular}

    \caption{Prompt used for temporal-guided clinical summarization
    (Part 1 of 3).}
    \label{tab:appendix_stage3_prompt}
\end{table*}

\begin{table*}[!htbp]
    \ContinuedFloat
    \centering
    \small
    \setlength{\tabcolsep}{5pt}
    \renewcommand{\arraystretch}{1.05}

    \begin{tabular}{p{0.10\linewidth} | p{0.85\linewidth}}
    \toprule
    \textbf{Type} & \textbf{Prompt}\\
    \toprule

    \textbf{System Prompt}
    &
    \textbf{INCLUDE}\\

    &Include only the following categories. If an event does not fall
    into one of these categories, omit it.\\

    &1. Patient demographics: date of birth, age, and sex.\\

    &2. The reason for admission: the clinical problem or finding that
    led to the admission, and any diagnostic workup or counseling
    directly related to it.\\

    &3. In-stay events, limited to the following types:\\

    &\hspace{1em}a. Procedures and surgeries performed.\\
    &\hspace{1em}b. Intraoperative findings and repairs.\\
    &\hspace{1em}c. Admissions to and transfers between services or
    units.\\
    &\hspace{1em}d. Intubation, extubation, and other airway or
    ventilation events.\\
    &\hspace{1em}e. Placement or removal of devices, lines, tubes, or
    catheters.\\
    &\hspace{1em}f. Treatments or interventions started or stopped.\\
    &\hspace{1em}g. Diagnoses made or revised during the stay.\\
    &\hspace{1em}h. Consultations obtained.\\

    &4. The discharge: date of discharge, disposition, discharge
    condition, and follow-up instructions.\\

    &5. Social history relevant to the patient's mental health, unless it is described as non-contributory or
    insignificant in the source material.\\

    &6. Medications on admission and medications at discharge must
    each be rendered as a line-separated list under their own temporal
    prefix, with one medication per line copied verbatim from the event
    text. These are the only lists permitted in the output; all other
    content must be continuous prose, as in the following example:\\

    &\texttt{Before admission, the patient was taking:}\\
    &\hspace{1em}\textbullet\ \texttt{Medicine A at X dosage}\\
    &\hspace{1em}\textbullet\ \texttt{Medicine B}\\

    &\texttt{Approximately on [discharge date], the patient was
    discharged on:}\\
    &\hspace{1em}\textbullet\ \texttt{Medicine C for Y duration}\\
    &\hspace{1em}\textbullet\ \texttt{Medicine D via route Z}\\

    &\\[-4pt]
    &\textbf{EXCLUDE}\\

    &Exclude the following categories entirely:\\

    &1. Laboratory values and laboratory results.\\

    &2. Vital-sign values.\\

    &3. Physical-examination findings, including normal and benign
    findings and descriptions of body habitus or affect.\\

    &4. Statements that a parameter or status remained stable,
    unchanged, or within normal limits.\\

    &5. Past medical history, past surgical history, and prior events
    that are not the reason for the current admission, unless the
    condition actively complicated the hospital course, required a
    consultation, or significantly altered medical management during
    the stay.\\

    &6. Allergies.\\

    &\\[-4pt]
    &\textbf{FIDELITY}\\

    &\textbullet\ Use only facts present in the tagged event list.\\

    &\textbullet\ Do not infer, interpret, diagnose, or add clinical
    judgment.\\

    &\textbullet\ Do not alter any numeric value, date, medication
    name, dose, route, or frequency.\\

    &\textbullet\ Preserve negations exactly. ``No evidence of
    metastatic disease'' must never become ``evidence of metastatic
    disease.''\\

    &\textbullet\ Retain de-identification placeholders, such as
    \texttt{[Last Name (STitle) 9125]}, exactly as written. Do not
    expand, guess, or replace them.\\

    &\textbullet\ Do not add a heading, preamble, commentary, or
    closing statement.\\

    \bottomrule
    \end{tabular}

    \caption[]{Prompt used for temporal-guided clinical summarization
    (Part 2 of 3).}
\end{table*}

\begin{table*}[!htbp]
    \ContinuedFloat
    \centering
    \small
    \setlength{\tabcolsep}{5pt}
    \renewcommand{\arraystretch}{1.05}

    \begin{tabular}{p{0.10\linewidth} | p{0.85\linewidth}}
    \toprule
    \textbf{Type} & \textbf{Prompt}\\
    \toprule

    \textbf{System Prompt}
    &
    \textbf{ORDER AND FORMAT}\\

    &\textbullet\ Output continuous prose, except for the two
    medication lists specified above. Do not use other bullet points,
    headings, or tables.\\

    &\textbullet\ Open with the demographic sentence.\\

    &\textbullet\ Order all remaining content chronologically by the
    start date of its tag. For events sharing a start date, preserve
    the order in which they appear in the tagged event list.\\

    &\textbullet\ Place \texttt{[PRE ADM]} events, which carry no date,
    together as one ``Before admission'' group immediately before the
    admission date.\\

    &\textbullet\ Start a new paragraph at each of the following
    points: before admission, the admission date, and the discharge
    date.\\

    &\\[-4pt]
    &\textbf{WORKED CONVERSIONS}\\

    &These examples demonstrate temporal rendering only. Whether an
    event is retained is determined by the \textbf{INCLUDE} and
    \textbf{EXCLUDE} rules above.\\

    &\textbf{Input:}
    \texttt{- [2000-01-01] [EXACT] The patient was admitted on
    2000-01-01.}\\
    &\textbf{Output:}
    On 2000-01-01, the patient was admitted.\\

    &\\[-4pt]
    &\textbf{Input:}
    \texttt{- [2000-01-03] [APPROX] She was extubated on postoperative
    day three.}\\
    &\textbf{Output:}
    Approximately on 2000-01-03, she was extubated.\\

    &\\[-4pt]
    &\textbf{Input:}
    \texttt{- [2000-05-10 to 2000-05-30] [APPROX] The patient was
    admitted to the ICU service.}\\
    &\textbf{Output:}
    Approximately from 2000-05-10 to 2000-05-30, the patient was
    admitted to the ICU service.\\

    &\\[-4pt]
    &\textbf{Input:}
    \texttt{- [PRE ADM] The patient was found to be unconscious.}\\
    &\textbf{Output:}
    Before admission, the patient was found to be unconscious.\\

    &\\[-4pt]
    &\textbf{Input:}
    \texttt{- [2011] [EXACT] The patient is status post knee operation
    in 2011.}\\
    &\textbf{Output:}
    In 2011, the patient underwent knee operation.\\

    &\\[-4pt]
    &\textbf{Merging two events sharing one tag:}\\

    &\textbf{Input:}
    \texttt{- [2013-12-31] [APPROX] The patient's social service
    worker recommended therapy.}\\

    &\texttt{- [2013-12-31] [APPROX] The patient's social service
    worker discussed treatment plans with the patient's PCP.}\\

    &\textbf{Output:}
    Approximately on 2013-12-31, the patient's social service worker
    recommended therapy and discussed treatment plans with the
    patient's PCP.\\

    &\\[-4pt]
    &\textbf{Subsumption:}\\

    &\textbf{Input:}
    \texttt{- [PRE ADM] The patient is female.}\\

    &\texttt{- [PRE ADM] The patient is a 58-year-old female.}\\

    &\textbf{Output:}
    Keep only the second event and incorporate it into the opening
    demographic sentence.\\

    &\\[-4pt]
    &Now produce the summary.\\

    &\\
    \midrule

    \textbf{User Prompt}
    &
    ORIGINAL\_DISCHARGE\_SUMMARY:\\

    &\texttt{\textless discharge\_summary\textgreater}\\

    &\\[-4pt]
    &TAGGED\_EVENT\_LIST:\\

    &\texttt{\textless tagged\_event\_list\textgreater}\\

    \bottomrule
    \end{tabular}

    \caption[]{Prompt used for temporal-guided clinical summarization
    (Part 3 of 3).}
\end{table*}

\section{Stage 3 Evaluation Rubrics}
\label{app:stage3_eval_rubrics}
Stage 3 is evaluated by both human annotators and the calibrated LLM evaluator using two shared dimensions. Faithfulness to Source measures whether each clinical claim in the generated summary is supported by the human-verified event timeline and preserves its original clinical meaning. Temporal Accuracy measures whether included events are rendered under the correct date and certainty tag and whether only events with identical temporal tags are grouped together. The human-verified timeline serves as the reference of record; the original discharge summary is used only to clarify unclear wording. Because selective content inclusion is expected in summarization, omitted events are not penalized under either dimension. The complete rubrics are provided in Tables~\ref{tab:appendix_stage3_faithfulness} and
\ref{tab:appendix_stage3_temporal}.

\begin{table*}[!htbp]
    \centering
    \small
    \setlength{\tabcolsep}{5pt}
    \renewcommand{\arraystretch}{1.08}

    \begin{tabular}{p{0.16\linewidth} | p{0.79\linewidth}}
        \toprule
        \textbf{Component}
        & \textbf{Dimension A: Faithfulness to Source} \\
        \toprule

        \textbf{Guiding Question}
        &
        Is every clinical event and fact in the summary traceable to the
        source discharge summary or the human-verified event timeline,
        with no invented content, and is its clinical meaning preserved?
        Temporal placement is evaluated separately under Dimension B.
        \\

        &\\[-4pt]

        \textbf{Reference}
        &
        Evaluate the candidate summary primarily against the human-verified
        event timeline. The source discharge summary may be used only to
        clarify an event's wording when the timeline is unclear. Do not
        reward or penalize content that appears in the source note but is
        absent from the verified timeline.
        \\

        &\\[-4pt]

        \textbf{Faithful Claim}
        &
        A claim is faithful when it is traceable to a corresponding event
        and preserves the event's clinical meaning. Permitted rewriting,
        such as expanding an abbreviation while retaining the original or
        supplying a subject and verb for structured content, is not an
        error when the meaning remains unchanged. The summary must include all mental health-related and social history events from the original source.
        \\

        &\\[-4pt]

        \textbf{Untraceable Claims}
        &
        A claim is unfaithful when no supporting event exists in the
        human-verified timeline or source note. This includes an invented
        event, fact, medication, diagnosis, or clinical detail.
        \\

        &\\[-4pt]

        \textbf{Meaning Distortions}
        &
        A claim is also unfaithful when it corresponds to a real event but
        changes its clinical meaning. Examples include:
        \\

        &
        \textbullet\ adding, dropping, or reversing a negation;
        \\

        &
        \textbullet\ adding or removing a hedge or uncertainty marker;
        \\

        &
        \textbullet\ changing a numeric value, dose, medication name,
        route, frequency, or laterality;
        \\

        &
        \textbullet\ introducing an unsupported causal explanation,
        diagnosis, or clinical rationale; or
        \\

        &
        \textbullet\ merging events into a new claim that neither source
        event supports.
        \\

        &\\[-4pt]

        \textbf{Do Not Assess}
        &
        Do not penalize omitted events, because selective inclusion is
        expected in a summary. Do not assess wrong dates, temporal tags,
        or temporal grouping under this dimension; these are evaluated
        under Dimension B.
        \\

        &\\[-4pt]

        \textbf{Rate}
        &
        The faithfulness rate is the percentage of clinical claims in the
        summary that are both traceable to the reference and
        meaning-preserving.
        \\

        \midrule

        \textbf{Score 5}
        &
        At least 95\% of the summary's clinical claims are faithful.
        \\

        \textbf{Score 4}
        &
        At least 80\% but less than 95\% of the summary's clinical claims
        are faithful.
        \\

        \textbf{Score 3}
        &
        At least 60\% but less than 80\% of the summary's clinical claims
        are faithful.
        \\

        \textbf{Score 2}
        &
        At least 40\% but less than 60\% of the summary's clinical claims
        are faithful.
        \\

        \textbf{Score 1}
        &
        Less than 40\% of the summary's clinical claims are faithful.
        \\

        \bottomrule
    \end{tabular}

    \caption{Stage-3 evaluation rubric for Dimension A:
    Faithfulness to Source.}
    \label{tab:appendix_stage3_faithfulness}
\end{table*}

\begin{table*}[!htbp]
    \centering
    \small
    \setlength{\tabcolsep}{5pt}
    \renewcommand{\arraystretch}{1.08}

    \begin{tabular}{p{0.16\linewidth} | p{0.79\linewidth}}
        \toprule
        \textbf{Component}
        & \textbf{Dimension B: Temporal Accuracy} \\
        \toprule

        \textbf{Guiding Question}
        &
        Is every clinical claim in the summary placed under the same
        temporal tag assigned in the human-verified event timeline, and
        are only events with identical tags grouped together?
        \\

        &\\[-4pt]

        \textbf{Scope}
        &
        Evaluate temporal placement and grouping only. Whether an event is
        supported by the reference and whether its clinical meaning is
        preserved are assessed under Dimension A.
        \\

        &\\[-4pt]

        \textbf{Temporal Tags}
        &
        Every event carries one of the following temporal tags:
        \\

        &
        \textbullet\ \texttt{[DATE] [EXACT]}, where \texttt{DATE} is a
        single ISO date or ISO date range;
        \\

        &
        \textbullet\ \texttt{[DATE] [APPROX]}, where \texttt{DATE} is a
        single ISO date or ISO date range; or
        \\

        &
        \textbullet\ \texttt{[PRE ADM]}, for an event occurring before
        admission without a resolved date.
        \\

        &\\[-4pt]

        \textbf{Valid Grouping}
        &
        A grouping is valid only when all events within it share an
        identical temporal tag: the same date or date range and the same
        certainty marker. Events with the same date but different
        \texttt{[EXACT]} and \texttt{[APPROX]} markers must not be merged.
        \\

        &\\[-4pt]

        \textbf{Misplacement}
        &
        A claim is temporally misplaced when it appears under a different
        date, date range, or certainty tag from the one assigned in the
        verified timeline. Examples include placing a discharge-day event
        on the admission date, rendering an \texttt{[EXACT]} event as a
        date range, or assigning a date to a \texttt{[PRE ADM]} event.
        \\

        &\\[-4pt]

        \textbf{Invalid Grouping}
        &
        A temporal grouping is incorrect when:
        \\

        &
        \textbullet\ events with different dates or date ranges are merged;
        \\

        &
        \textbullet\ \texttt{[EXACT]} and \texttt{[APPROX]} events are
        merged despite sharing the same date;
        \\

        &
        \textbullet\ a \texttt{[PRE ADM]} event is merged with a dated
        event; or
        \\

        &
        \textbullet\ events sharing one identical tag are unnecessarily
        split across multiple temporal prefixes.
        \\

        &\\[-4pt]

        \textbf{Do Not Assess}
        &
        Do not assess invented or meaning-distorted events under this
        dimension; these are evaluated under Dimension A. Do not penalize
        omitted events. The order of correctly tagged groups is not an
        error, provided each group appears under its correct temporal
        prefix.
        \\

        &\\[-4pt]

        \textbf{Rate}
        &
        The temporal-accuracy rate is the percentage of included clinical
        events that are both correctly placed and correctly grouped.
        Rates are calculated to one decimal place, without rounding before
        assigning the score band. The counts used to calculate the rate
        must also be reported.
        \\

        \midrule

        \textbf{Score 5}
        &
        At least 95\% of included events are correctly placed and grouped.
        \\

        \textbf{Score 4}
        &
        At least 80\% but less than 95\% of included events are correctly
        placed and grouped.
        \\

        \textbf{Score 3}
        &
        At least 60\% but less than 80\% of included events are correctly
        placed and grouped.
        \\

        \textbf{Score 2}
        &
        At least 40\% but less than 60\% of included events are correctly
        placed and grouped.
        \\

        \textbf{Score 1}
        &
        Less than 40\% of included events are correctly placed and grouped.
        \\

        \bottomrule
    \end{tabular}

    \caption{Stage-3 evaluation rubric for Dimension B:
    Temporal Accuracy.}
    \label{tab:appendix_stage3_temporal}
\end{table*}

\section{Stage 3 LLM and Auto-Evaluator Results}
\label{app:stage3_llm_auto_eval_results}

\begin{table*}[t]
    \centering
    \small
    \setlength{\tabcolsep}{8pt}
    \renewcommand{\arraystretch}{1.15}

    \begin{tabular}{@{}lccccc@{}}
        \toprule
        \textbf{Model}
        & \textbf{Faithfulness}
        & \textbf{Temporal Accuracy}
        & \multicolumn{3}{c}{\textbf{MEDCON}} \\
        \cmidrule(lr){4-6}

        \textbf{(summaries)}
        & \textbf{Band / \%}
        & \textbf{Band / \%}
        & \textbf{P}
        & \textbf{R}
        & \textbf{F1} \\
        \midrule

        Gemini 2.5 Pro
        & 5.00 / 99.93
        & 4.75 / 95.33
        & 0.725
        & 0.519
        & 0.599 \\
        \bottomrule
    \end{tabular}

    \caption{Evaluation results for summaries generated by Gemini 2.5 Pro. Faithfulness and temporal accuracy are reported as mean rubric bands and percentage scores. MEDCON precision, recall, and F1 measure clinical-concept overlap.}
    \label{tab:stage3_summary_results}
\end{table*}

As shown in Table~\ref{tab:stage3_summary_results}, faithfulness is saturated: all 52 summaries score band 5, with 5 distorted claims overall. 

Temporal accuracy shows the mean likert band at 4.75 / 5, macro-mean 95.33\% of events correctly placed (micro 96.17\%, 257 errors overall; median 100\%).

MEDCON precision ($0.725$) exceeds recall ($0.519$), F1 $0.599$. Asserted content is largely grounded, consistent with the faithfulness band; roughly half of source concepts are omitted. The judge scores only claims that were made and is therefore blind to this coverage gap, while MEDCON is insensitive to temporal placement. The two measure disjoint failure axes and are reported together for that reason.

\section{Stage 3 Evaluation Agreement}
\label{app:stage3_eval_agreement}
\subsection{Annotation Protocol and Unit Definitions}

\begin{table*}[!t]
\centering
\small
\setlength{\tabcolsep}{3.8pt}
\renewcommand{\arraystretch}{1.10}
\resizebox{0.95\textwidth}{!}{%
\begin{tabular}{@{}llccccccccc@{}}
\toprule
& & \multicolumn{3}{c}{\textsc{Silver$\rightarrow$Silver}}
& \multicolumn{3}{c}{\textsc{Silver$\rightarrow$Gold}}
& \multicolumn{3}{c}{\textsc{Gold}} \\
\cmidrule(lr){3-5}
\cmidrule(lr){6-8}
\cmidrule(lr){9-11}

\textbf{Metric}
& \textbf{Model}
& \textbf{Zero} & \textbf{Few} & \textbf{IT}
& \textbf{Zero} & \textbf{Few} & \textbf{IT}
& \textbf{Zero} & \textbf{Few} & \textbf{IT} \\
\midrule

\multirow{5}{*}{ROUGE-1}
& Gemma4-31B
& .826$\pm$.008 & .823$\pm$.008 & \textbf{.886$\pm$.006}
& .805$\pm$.011 & .794$\pm$.013 & .865$\pm$.013
& -- & \textbf{.743$\pm$.026} & .729$\pm$.030 \\

& Qwen3.6-35B-A3B
& .749$\pm$.010 & .791$\pm$.009 & .879$\pm$.007
& .717$\pm$.014 & .754$\pm$.016 & \textbf{.867$\pm$.011}
& -- & .670$\pm$.036 & .656$\pm$.037 \\

& Qwen3-30B-A3B
& .737$\pm$.013 & .750$\pm$.013 & .843$\pm$.012
& .713$\pm$.014 & .765$\pm$.012 & .811$\pm$.018
& -- & .136$\pm$.040 & .296$\pm$.041 \\

& Mistral-24B
& .736$\pm$.011 & .761$\pm$.011 & .878$\pm$.007
& .679$\pm$.021 & .724$\pm$.017 & .864$\pm$.011
& -- & .698$\pm$.022 & \textbf{.763$\pm$.027} \\

& Meditron-7B
& .555$\pm$.021 & .595$\pm$.022 & .804$\pm$.017
& .471$\pm$.030 & .531$\pm$.031 & .802$\pm$.021
& -- & .178$\pm$.043 & .230$\pm$.031 \\

\midrule

\multirow{5}{*}{ROUGE-2}
& Gemma4-31B
& .711$\pm$.010 & .693$\pm$.011 & \textbf{.804$\pm$.009}
& .675$\pm$.013 & .649$\pm$.017 & .776$\pm$.015
& -- & \textbf{.607$\pm$.024} & .629$\pm$.028 \\

& Qwen3.6-35B-A3B
& .621$\pm$.011 & .671$\pm$.011 & .797$\pm$.009
& .575$\pm$.016 & .622$\pm$.018 & \textbf{.782$\pm$.012}
& -- & .555$\pm$.032 & .551$\pm$.034 \\

& Qwen3-30B-A3B
& .598$\pm$.012 & .617$\pm$.013 & .748$\pm$.013
& .567$\pm$.015 & .613$\pm$.015 & .708$\pm$.019
& -- & .112$\pm$.033 & .229$\pm$.034 \\

& Mistral-24B
& .600$\pm$.012 & .640$\pm$.013 & .795$\pm$.009
& .540$\pm$.020 & .597$\pm$.019 & .778$\pm$.013
& -- & .564$\pm$.021 & \textbf{.655$\pm$.026} \\

& Meditron-7B
& .422$\pm$.019 & .463$\pm$.020 & .709$\pm$.018
& .336$\pm$.026 & .399$\pm$.027 & .696$\pm$.022
& -- & .142$\pm$.035 & .141$\pm$.025 \\

\bottomrule
\end{tabular}
}
\caption{
Additional lexical evaluation for Task~C timeline summarization.
We report ROUGE-1 and ROUGE-2 F1 under
\textsc{Silver$\rightarrow$Silver},
\textsc{Silver$\rightarrow$Gold}, and
\textsc{Gold}.
Results are mean $\pm$ bootstrap SD over 2,000 document-level resamples.
Higher is better.
}
\label{tab:task-c-additional-rouge}
\end{table*}

Human annotators reviewed all 52 benchmark summaries alongside the judge's output. The review interface presents the judge's findings as a list and allows the annotator to mark any finding not a mistake.




\begin{table}[t]
    \centering
    \small
    \setlength{\tabcolsep}{6pt}
    \renewcommand{\arraystretch}{1.10}

    \begin{tabular}{lcc}
        \toprule
        \textbf{Metric} & \textbf{Value} & \textbf{95\% CI} \\
        \midrule
        Observed agreement (\(P_o\))
        & 0.904
        & [0.808, 0.981] \\

        Cohen's \(\kappa\)
        & 0.773
        & [0.567, 0.947] \\

        Gwet's AC1
        & 0.835
        & [0.674, 0.966] \\

        Positive agreement
        & 0.839
        & [0.667, 0.960] \\

        Negative agreement
        & 0.932
        & [0.862, 0.986] \\
        \bottomrule
    \end{tabular}

    \vspace{2pt}
    \caption{Agreement between the automated evaluator and human annotations. \(P_o\) denotes observed agreement. Confidence intervals are
    reported at the 95\% level.}
    \label{tab:stage3_summary_agreement}
\end{table}

\subsection{Summary-level Detection Agreement}
Table~\ref{tab:stage3_summary_agreement} treats each summary as one unit and asks whether the judge and the annotator agree that it contains at least one error. This framing avoids committing to a claim inventory as the denominator, and at a positive rate of 35\% the prevalence index falls to 0.40, so Cohen's $\kappa$ and Gwet's AC1 no longer diverge and both are interpretable. The judge flagged 18 summaries; the annotator's adjudication left at least one surviving flag in 13 of them and eliminated all flags in 5.


\subsection{Error-count concordance}
\begin{table}[!h]
    \centering
    \small
    \setlength{\tabcolsep}{5pt}
    \renewcommand{\arraystretch}{1.10}

    \begin{tabular}{
        @{}
        p{0.50\columnwidth}
        p{0.18\columnwidth}
        p{0.23\columnwidth}
        @{}
    }
        \toprule
        \textbf{Metric} & \textbf{Value} & \textbf{95\% CI} \\
        \midrule

        \multicolumn{3}{@{}l}{\textit{Totals (\(n=52\) summaries)}} \\
        Errors flagged by judge
        & 262
        & -- \\

        Errors upheld by annotator
        & 39
        & -- \\

        Inflation ratio
        & \(6.7\times\)
        & -- \\

        \addlinespace
        \multicolumn{3}{@{}l}{\textit{Concordance}} \\
        Lin's CCC
        & 0.103
        & [0.031, 0.345] \\

        ICC(A,1)
        & 0.105
        & [0.032, 0.349] \\

        Pearson's \(r\)
        & 0.511
        & [0.119, 0.818] \\

        Spearman's \(\rho\)
        & 0.768
        & [0.573, 0.934] \\

        \addlinespace
        \multicolumn{3}{@{}l}{\textit{Bland--Altman analysis}} \\
        Bias (judge -- annotator)
        & \(+4.29\)
        & SD \(=15.32\) \\

        Limits of agreement
        & \multicolumn{2}{l}{\(-25.7\) to \(+34.3\)} \\

        \bottomrule
    \end{tabular}

    \vspace{2pt}
    \caption{Error-count concordance between the automated judge and human annotator. Per-summary agreement between the judge's error count and the
    number of errors upheld by the annotator. CCC denotes Lin's concordance correlation coefficient; ICC(A,1) denotes the absolute-agreement intraclass correlation coefficient.}
    \label{tab:stage3_error_count_concordance}
\end{table}

Table~\ref{tab:stage3_error_count_concordance} compares the number of errors the judge reports in each summary against the number the annotator upheld. The two disagree sharply in magnitude and agree well in order. Lin's concordance correlation coefficient is 0.103 and ICC(A,1) is 0.105, both near the floor, because the judge over-reports by a mean of 4.3 errors per summary with 95\% limits of agreement spanning $-25.7$ to $+34.3$ -- an interval wider than the error count of any but the most heavily flagged summary. Spearman's $\rho$ is 0.768, however: the judge ranks summaries by error burden much as the annotator does.

The gap between Spearman's $\rho$ (0.768) and Pearson's $r$ (0.511) reflects a single influential summary, subject 7676, at 100 judge errors against 5 upheld; rank-based statistics absorb it and moment-based ones do not. Taken together these results characterize the judge as a triage and ranking instrument rather than a measurement one. Its absolute error counts should not be reported as validated error rates, and any downstream comparison between generation models should use the judge's ordering rather than its magnitudes.

\section{Stage 3 Evaluation Platform}
\label{app:stage3_eval_platform}
For Stage 3, we built a self-contained, locally-run web application for human review and correction of Gemini 2.5 Pro-generated Stage 3 summaries.

\begin{figure*}[t]
    \centering
    \includegraphics[width=\textwidth]{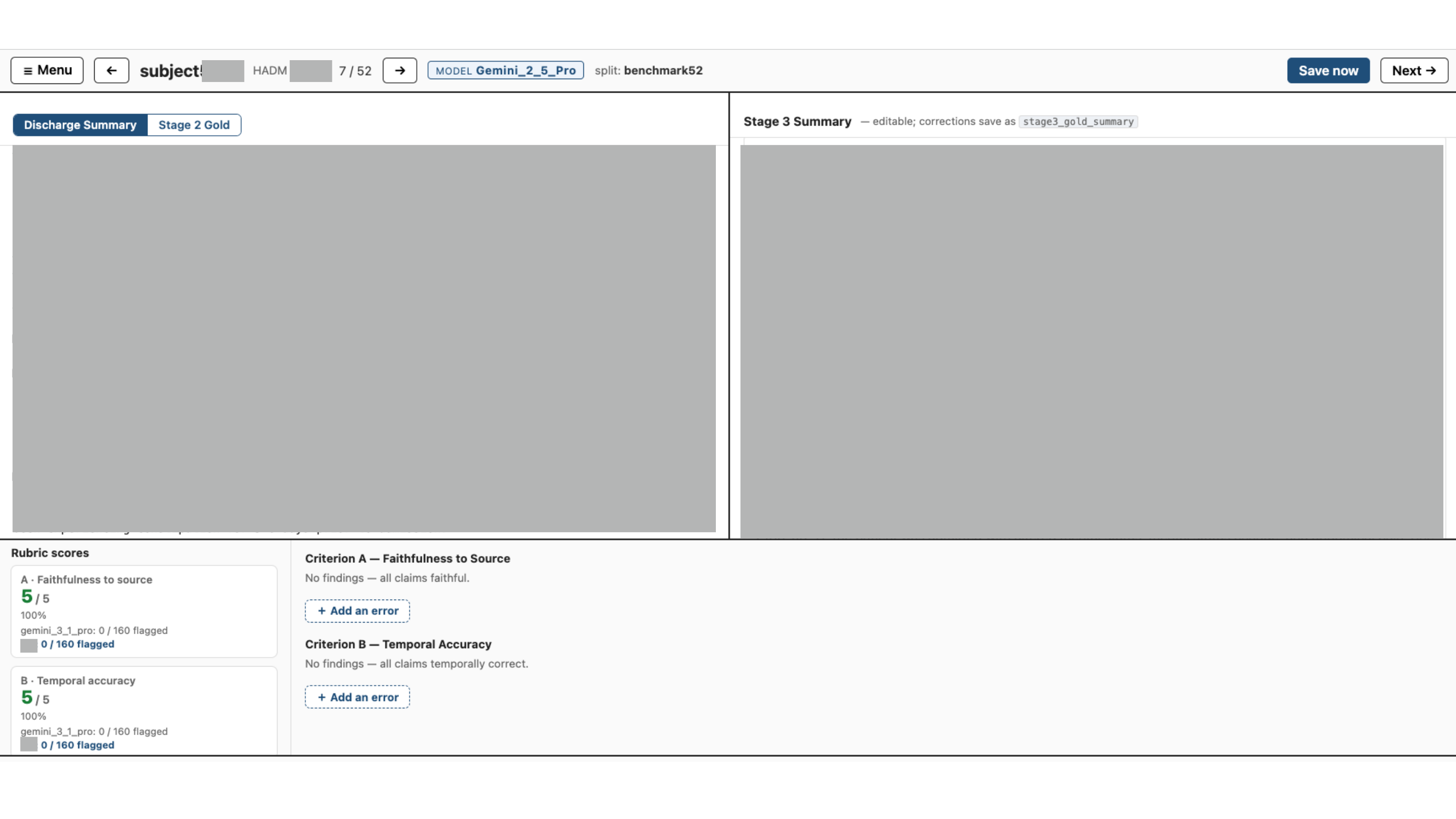}
    \caption{The human review and correction platform. For each subject, the reviewer compares the model's Stage 3 summary against the source and corrects it in place. Left: source panel, toggled between the discharge summary and Stage 2 gold events. Right: the model's Stage 3 summary in an editable field, saved as a separate field. Patient-text regions from the original MIMIC III source materials are redacted (grey).}
    \label{fig:appendix_stage3_annotation_platform}
\end{figure*}

The review screen has three regions (Figure~\ref{fig:appendix_stage3_annotation_platform}). The left panel shows the source material, toggling between the original discharge summary and the Stage 2 gold event list, and is read-only. The right panel presents the model's Stage 3 summary in an editable field: rather than only flagging problems, the reviewer corrects the summary in place to produce a gold summary. Edits autosave, and an "edited" indicator with a revert-to-original control distinguishes corrected from untouched summaries; the correction is stored as a separate \texttt{stage3\_gold\_summary} field, preserving the model's original output for comparison. The bottom panel surfaces an LLM-judge evaluation as a scaffold for verification. The summary is scored on two rubric criteria as listed in Appendix~\ref{app:stage3_eval_rubrics}. 

\section{Qualitative Analysis Cases}
\label{app:qualitative_analysis_cases}
In this section, we analyze three clinician-verified gold cases using WHO categories of mental disorders, symptoms, and risk factors as mentioned in Section~\ref{sec:qualitative_analysis}, grounding each assignment in the gold timeline and treating unestablished links as co-documented rather than causal.

\subsection{Case A: Chronic Disability and an In-Stay Depression}
\label{sec:qualitative_analysis_case_A}
Case A comprises 306 gold events across a 3-week admission. The record describes roughly three decades of C5 quadriplegia, an Individual-level risk factor that Stage~2 resolves to a year-only \texttt{[APPROX]} date some thirty years before admission, and a marked deterioration in ventilatory status during the stay. The note itself states the sequence, attributing the patient's depressed mood to that deterioration, so the link is explicit rather than inferred by us. Depression is documented as a diagnosis: pharmacotherapy for situational depression is started during the admission, and psychiatry is consulted on the 11th hospital day. The symptoms recorded are suicidal thoughts and refusal of food; at discharge the note records restored hope, concentration and sleep, and continuation of an SSRI. Several symptoms are therefore attested only through their documented reversal, an implicit form of expression that an extraction system keyed to symptom mentions alone would miss.
 
What makes this case instructive is where the onset falls. The background risk factor begins decades before admission, whereas the depressive episode emerges during the hospital stay, making their temporal relationship difficult to recover from the narrative alone. Only a calendar-anchored representation places them on one ordered chain while preserving that separation. The distinction is also fragile in exactly the way our error analysis predicts: had the onset been tagged \texttt{[PRE ADM]} rather than resolved to the admission interval, the clinically salient fact that this depression arose during the stay would disappear from the timeline entirely. This is the same boundary that accounts for the largest class of temporal disagreements in Section~\ref{sec:stage1_stage2_eval_agreement}.

\subsection{Case B: Depression Across All Three WHO Risk Levels}
\label{sec:qualitative_analysis_case_B}
Case B contains 494 gold events spanning a 1.5-month admission after an intentional overdose. It is the only case in our sample with documented risk factors across all three WHO levels. At the individual level, the record describes long-standing alcohol misuse across the past medical history, social history, and discharge diagnoses. At the Family \& Community level, it notes limited social support, poor self-care, and the recent loss of a living arrangement with a relative. At the Structural level, it identifies multiple social stressors and a lack of insurance coverage. These links are stated directly in the record, which cites them as reasons for involving psychiatry and social work. The associated symptoms and behaviors include the presenting overdose, two earlier suicide attempts by different means, a discharge diagnosis noting multiple attempts, and a documented reduction in suicidal intent during the admission.

This case highlights two important features of the reconstruction. First, the structural risk factors are exactly the kind of information that a fixed clinical schema might omit, even though they are central to the explanation recorded by the care team. This illustrates the need for the open-vocabulary representation described in Section~\ref{sec:introduction}. Second, the record contains an unresolved contradiction: psychiatry documents that the patient was not currently depressed during the admission, while depression remains listed among the discharge diagnoses. 

\subsection{Case C: A Risk Cascade Under an Unsettled Diagnosis}
\label{sec:qualitative_analysis_case_C}

Case C contains 227 gold events spanning a 12-day admission and includes four documented risk factors, only one of which the source explicitly links to the diagnosis. The record identifies more than two decades of alcohol misuse as clinically relevant to the patient’s diagnosis. It also documents childhood sexual abuse by a caregiver, job loss attributed to drinking, and living alone after the end of a relationship about a year before admission. Because the note does not connect these latter factors directly to the disorder, we treat them as co-documented rather than causal. Bipolar disorder appears as an established diagnosis, supported by mood-stabilizing treatment at admission, repeated overdose-related hospitalizations, a prior psychiatric admission after a suicide attempt, and multiple presentations following suicide threats.

What distinguishes this case is that the record later revises its own clinical interpretation. During the admission, psychiatry questions whether the bipolar diagnosis is accurate and considers whether long-term alcohol use may have contributed to a misdiagnosis. Most background risk factors are undated and therefore tagged \texttt{[PRE ADM]}, whereas this diagnostic reconsideration is placed within the admission interval. Temporal anchoring makes the progression visible: long-standing background factors, an established diagnosis, and then in-stay doubt about that diagnosis. The case also shows that uncertainty is not only temporal; it also concerns causality and diagnosis. Finally, it demonstrates the limits of surface matching. During screening, a statement about episodic drinking was incorrectly matched to an eating-disorder symptom term and was removed after clinician review, illustrating why lexical overlap alone cannot support reliable qualitative interpretation.

\section{Additional Results for Evaluation}
\label{app:additional-results}
We present additional evaluation results for the open-weight models (Table~\ref{tab:task-c-additional-rouge}).
\end{document}